\documentclass{article}

\usepackage{mathtools}
\usepackage[final]{neurips_2022}

\usepackage[utf8]{inputenc} % allow utf-8 input
\usepackage[T1]{fontenc}    % use 8-bit T1 fonts
\usepackage{hyperref}       % hyperlinks
\usepackage{url}            % simple URL typesetting
\usepackage{booktabs}       % professional-quality tables
\usepackage{amsfonts}       % blackboard math symbols
\usepackage{nicefrac}       % compact symbols for 1/2, etc.
\usepackage{microtype}      % microtypography
\usepackage{xcolor}         % colors
\usepackage{amsmath,amssymb,amsthm,amsfonts}
\usepackage{latexsym,bbm,graphicx,float,mathtools}
\usepackage{algorithmic}
\usepackage[export]{adjustbox}
\usepackage{capt-of}% or \usepackage{caption}
\usepackage{booktabs}
\usepackage{varwidth}
\newsavebox\tmpbox

\usepackage[noabbrev]{cleveref}
\usepackage{caption}
\usepackage{subcaption}
\usepackage{transparent}
\usepackage[inline]{enumitem}
\usepackage{multirow}
\usepackage{multicol}
\usepackage{algorithm}
\usepackage{xspace}

\newtheorem{thm}{Theorem}[section]

\newtheorem{lem}[thm]{Lemma}

\theoremstyle{newDef}

\numberwithin{equation}{section}

\newcommand\algname{\texttt{AQ-SGD}\xspace}

\newenvironment{proofof}[1]{\indent{\scshape Proof of #1}:~~}{\vspace{3mm}\qed}
\newenvironment{proofcap}{\indent{\scshape Proof}:~~}{\vspace{3mm}\qed}

\newcommand{\E}{\mathbb{E}}
\newcommand{\cD}{\mathcal{D}}
\newcommand{\inn}[2]{\langle #1, #2 \rangle}

\title{Fine-tuning Language Models over Slow Networks
using Activation Quantization
with Guarantees}

\author{
    Jue Wang$^{1,3}$\thanks{Equal contribution.\hspace{5mm}$^{\dagger}$Now at Google.}, Binhang Yuan$^{1}$\footnotemark[1], Luka Rimanic$^{1\dagger}$\footnotemark[1], Yongjun He$^1$, Tri Dao$^2$,\\
    \textbf{Beidi Chen}$^{34}$, \textbf{Christopher Ré}$^2$, \textbf{Ce Zhang}$^1$\\
    $^1$ETH Z\"urich, Switzerland~~~$^2$Stanford University, USA~~~$^3$Zhejiang University, China\\
    $^4$Carnegie Mellon University~~~$^5$Meta AI\\
\{juewang, binhang.yuan, luka.rimanic, yongjun.he, ce.zhang\}@inf.ethz.ch\\
\{beidic, trid, chrismre\}@stanford.edu
}

\begin{document}

\maketitle

\begin{abstract}
Communication compression is a crucial 
technique for modern distributed learning 
systems to alleviate their communication 
bottlenecks over slower networks.
Despite recent intensive studies
of gradient compression for data parallel-style
training, compressing the \textit{activations}
for models trained with
pipeline parallelism is still an
open problem. In this paper, we propose 
\algname, a novel activation compression 
algorithm for communication-efficient 
pipeline parallelism training 
over slow networks. Different from previous 
efforts in activation compression,
instead of compressing activation values directly, \algname compresses the \textit{changes of the
activations}. This allows us to show,
to the best of our knowledge for the first time, 
that one can still achieve 
$O(1/\sqrt{T})$ convergence rate for 
non-convex objectives
under 
activation compression, without making 
assumptions on gradient unbiasedness
that do not hold for deep learning models with non-linear activation functions.
We then show that \algname can be optimized
and implemented efficiently, without 
additional end-to-end runtime overhead.
We evaluated \algname to fine-tune
language models with up to 1.5 billion parameters,
compressing activation to 2-4 bits.
\algname provides up to $4.3\times$ end-to-end speed-up in slower networks, without
sacrificing model quality.
Moreover, we also show that \algname
can be combined with state-of-the-art 
gradient compression algorithms to enable 
``end-to-end communication compression'': \textit{All communications between machines, including model gradients, forward activations,
and backward gradients are compressed into lower precision}.
This provides up to $4.9\times$ end-to-end speed-up, without
sacrificing model quality.
\end{abstract}

\section{Introduction}
% \vspace{-1em}

{%\color{blue}
Decentralized or open collaborative training has recently attracted intensive interests~\cite{ryabinin2020towards,diskin2021distributed,borzunov2022training,ryabinin2021swarm}.
Despite their great potential in leveraging geo-distributed powerful GPUs, the computation efficiency is severely hindered by low network bandwidth --- typically in the range of 10-400Mbps \cite{diskin2021distributed,borzunov2022training,ryabinin2021swarm}.
} Recently, efforts in improving communication efficiency 
% for distributed learning 
have significantly decreased the dependency 
% of training deep learning models 
on fast data center networks --- 
the \textit{gradient} can be compressed to lower
precision or sparsified~\cite{alistarh2016qsgd,zhang2017zipml,bernstein2018signsgd,wen2017terngrad}, which speeds up training
over low bandwidth networks, whereas
the \textit{communication topology} can be decentralized
~\cite{koloskova2019decentralized,li2018pipe,lian2017can,lian2018asynchronous,tang2018communication,tang2018d}, which
speeds up training over high latency networks.
Indeed, today's state-of-the-art training systems, such as Pytorch~\cite{li13pytorch, pytorchlightning}, Horovod~\cite{sergeev2018horovod},
Bagua~\cite{gan2021bagua},
and
BytePS~\cite{jiang2020unified},
already support many of these communication-efficient training paradigms.

However, with the rise of large foundation models~\cite{DBLP:journals/corr/abs-2108-07258} (e.g., BERT \cite{devlin2018bert}, GPT-3 \cite{brown2020language}, and CLIP\cite{radford2021learning}),
improving communication efficiency via compression becomes 
more challenging.
Current training systems for foundation models such as Megatron~\cite{shoeybi2019megatron}, Deepspeed~\cite{rasley2020deepspeed}, and Fairscale~\cite{baines2021fairscale},
allocate different layers of the model onto multiple devices
and need to communicate ---
\textit{in addition to} the gradients on the models ---
the \textit{activations} during the
forward pass and the \textit{gradients
on the activations} during 
the backward pass. 
Compressing these 
\textit{activations} leads to a very different
behavior compared with compressing
the gradient --- \textit{simply compressing these
activations in a stochastically unbiased 
way will lead to biases in the gradient
that cannot be measured easily or 
expressed in closed form}.
This either breaks the unbiasedness assumptions
made by most gradient compression 
results~\cite{alistarh2016qsgd,zhang2017zipml,bernstein2018signsgd,wen2017terngrad}
or makes error compensation over
gradient biases~\cite{tang2019doublesqueeze,tang2019deepsqueeze}
difficult to adopt.

\begin{figure}[!t]
\begin{minipage}[t!]{\linewidth}
      \centering
      \begin{minipage}{0.61\linewidth}
        \begin{figure}[H]
        \centering
    \begin{subfigure}[b]{0.46\linewidth}
        \centering
        \includegraphics[width=\linewidth]{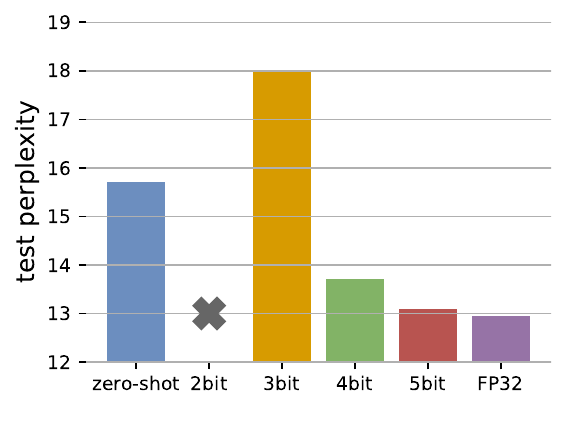}
        % \vspace{-6mm}
        \caption{Fine-tune WikiText2}
        \label{fig:intro_ppl}
    \end{subfigure}
    \begin{subfigure}[b]{0.46\linewidth}
        \centering
        \includegraphics[width=\linewidth]{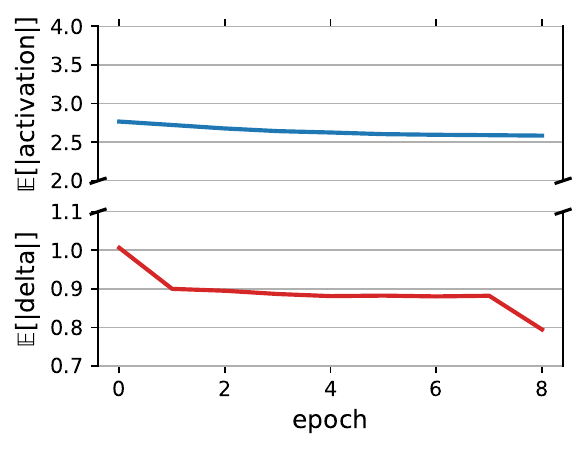}
        % \vspace{-6mm}
        \caption{Activation and delta}
        \label{fig:intro_vars}
    \end{subfigure}
    % \vspace{-2mm}
    \caption{(a) Fine-tuning GPT2-1.5B with different activation precisions in communication; 
            %  (b) distribution of activations for GPT2-1.5B during training; 
            %  (c) distribution of changes of activations for GPT2-1.5B during training. 
            (b) Average absolute value of activations and their changes for GPT2-1.5B during training.
            %  \luka{Shall we make these more visible (for example 1(c)? And more look-alike with Figure 3.} \jue{how about now?}
             }
    \label{fig:intro}
\end{figure}
      \end{minipage}
      \hspace{0.01\linewidth}
      \begin{minipage}{0.36\linewidth}
      % \vspace{-2mm}
        \begin{table}[H]
        \scriptsize
        \centering
         \caption{Summary of technical results. AC-GC~\cite{evans2021ac} and TinyScript~\cite{fu2020don} assume that the returned gradient is unbiased, whereas \algname algorithm does not rely on such an assumption.\vspace{0.08\linewidth}}
         \resizebox{\linewidth}{!}{
        \begin{tabular}{c|c|c}
        \hline
            Algorithm  & Assumptions & Conv. Rate   \\
             & on Quant. Grad. &  \\
        \hline
            SGD~\cite{bottou2018optimization}  & N/A    &   $\mathcal{O}(1/\sqrt{T})$  \\
        \hline
            AC-GC~\cite{evans2021ac} & Unbiased & $\mathcal{O}(1/\sqrt{T})$  \\
            TinyScript~\cite{fu2020don} & Unbiased & $\mathcal{O}(1/\sqrt{T})$  \\
        \hline
            \algname  & N/A   & $\mathcal{O}(1/\sqrt{T})$ \\
        \hline
        \end{tabular}
        \vspace{2mm}
       }
        \label{tab:my_label}
        \end{table}
      \end{minipage}
 \end{minipage}
%  \vspace{-2em}
 \end{figure}

Previous efforts on activation compression~\cite{han2016deep,hubara2017quantized,jain2018gist,chakrabarti2019backprop,chen2021actnn} illustrate, albeit mostly empirically, that 
large deep learning models can 
tolerate some compression errors on these activation values. 
However, when it comes to the underlying theoretical analysis, these efforts mostly 
make 
assumptions that do not apply to 
neural networks with non-linear activation functions --- the only two recent efforts that
claim theoretical convergence analysis~\cite{evans2021ac,fu2020don} assume that 
an unbiased compression on activations
leads to an unbiased error on the gradient.
Not surprisingly, these algorithms
lead to suboptimal quality
under relatively aggressive compression, illustrated in Figure~\ref{fig:intro_ppl} --- in many cases, using activation 
compression to fine-tune a model might 
be worse than  
zero-shot learning without any fine-tuning at all.

In this paper, we 
focus on 
the problem of activation compression for training language models over 
slow networks by asking the following:
\begin{itemize}
\item {\bf Q1.} {\em Can we design an algorithm for activation compression with rigorous 
theoretical guarantees on SGD convergence? }
\item {\bf Q2.} {\em Can such an algorithm 
be implemented efficiently without
additional runtime overhead and outperform today's activation compression algorithms
without accuracy loss? }
\end{itemize}

Our answers to both questions are \textbf{\textit{Yes}}. 
{\bf \underline{(Contribution 1)}} We propose \algname, a novel algorithm for activation compression. The idea of \algname is simple --- instead of directly compressing the activations,
\textit{compress the change of activations
for the same training example across epochs}.
Intuitively, we expect \algname to 
outperform simply compressing the activations
because it enables an interesting ``self-enforcing'' dynamics: 
\textit{the more training stabilizes $\rightarrow$
the smaller the changes of the model 
across epochs
$\rightarrow$
the smaller the changes of activations
for the same training example 
across epochs
$\rightarrow$
the smaller the compression error 
using the same \#bits
$\rightarrow$
training stabilizes more.}
{\bf \underline{(Contribution 2)}}
The theoretical analysis of \algname
is non-trivial since we have to analyze 
the above dynamics and connect it 
to SGD convergence, which is quite different
from most of today's results on 
gradient compression and error compensation.
Under mild technical conditions
and quantization functions with bounded error,
we show that 
\algname converges with a rate of 
$O(1/\sqrt{T})$ for non-convex 
objectives, the same as 
vanilla SGD~\cite{bottou2018optimization,liu2021distributed}.
To the best of our knowledge,
\algname is the first activation
compression algorithm with rigorous theoretical analysis that shows a convergence rate of $O(1/\sqrt{T})$ (without relying on assumptions of unbiased gradient). {\bf \underline{(Contribution 3)}}
We then show that \algname can be optimized and implemented efficiently\footnote{Our code is available at:  \url{https://github.com/DS3Lab/AC-SGD}.}, without adding 
additional end-to-end runtime overhead 
over non-compression and other compression schemes (it does require us to utilize more 
memory and SSD for storage of activations).
{\bf \underline{(Contribution 4)}} We then conduct extensive experiments on 
sequence classification and language modeling datasets using DeBERTa-1.5B and GPT2-1.5B models, respectively.
We show that \algname can aggressively quantize activations to 2-4 bits without sacrificing convergence performance, where direct quantization of activations fails to converge;
in slow networks, \algname achieves up to $4.3\times$ 
end-to-end speedup.
{\bf \underline{(Contribution 5)}}
Last but not least, we also show that \algname
can be combined with state-of-the-art 
gradient compression algorithms to enable 
``end-to-end communication compression'': \textit{All data exchanges between machines, including model gradients, forward activations,
and backward gradients are quantized into lower precision}.
This provides up to $4.9\times$ end-to-end speed-up, without
sacrificing model quality.

\begin{figure}[t!]
    \centering
    \includegraphics[width=0.9\linewidth]{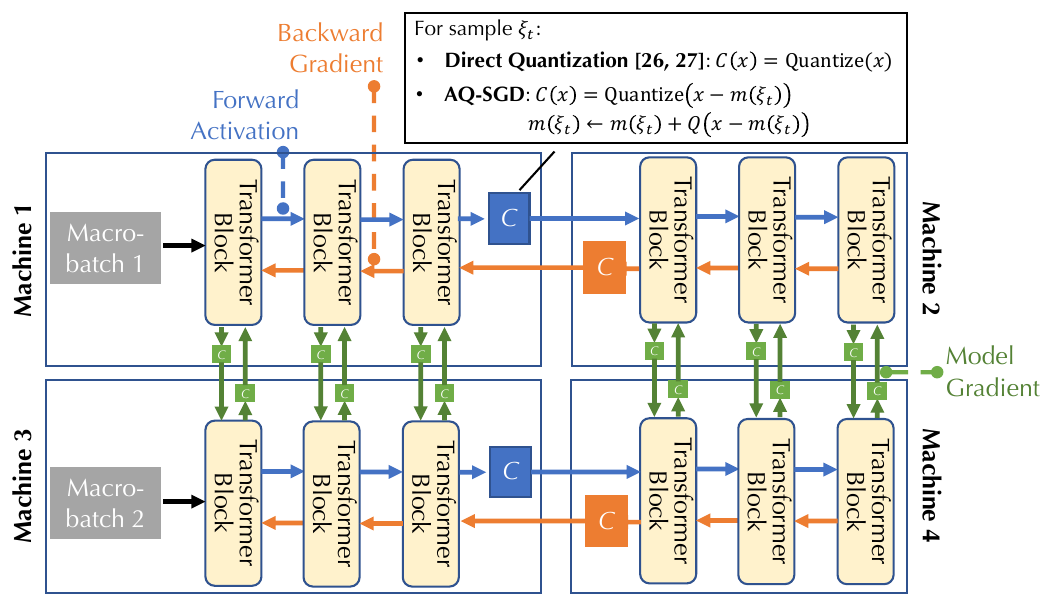}
    % \vspace{-0.5em}
    \caption{The communication pattern of training large language models with both 
    data parallelism and pipeline model parallelism. $C$ denotes a compression module. The goal of this paper is to understand the design of $C$ for \textit{forward activation} and \textit{backward gradient}.}
    \label{fig:illustration}
    % \vspace{-.5em}
\end{figure}

% \vspace{-0.5em}
\section{Overview and Problem Formulation}
% \vspace{-1em}
Training large language models over 
multiple devices is a challenging task. Because of the vast number of parameters of the model and data examples, state-of-the-art systems 
need to combine different forms of parallelism.
Figure~\ref{fig:illustration} illustrates 
an example in which such a model is 
distributed over four machines: 
{\em (Pipeline Model Parallelism)} 
The model is partitioned onto 
Machine 1 and Machine 2 (similarly, Machine 3 and Machine 4), each of which is holding
a subset of \textit{layers}.
To compute the gradient over the model
using backpropagation,
these two machines need to communicate the
\textit{activations} during the forward pass
and the \textit{gradient on activations}
during the backward pass.
{\em (Data Parallelism)} Machine 1 and Machine 3 (similarly, Machine 2 and Machine 4)
process the same set of \textit{layers}
for different \textit{macro-batches}.
In this case, each of them will hold a
replica of the same model. After the gradient 
over their model parameters are ready,
they need to communicate the 
\textit{model gradient}, usually 
by taking an average~\cite{li13pytorch,sergeev2018horovod,gan2021bagua}.

% \vspace{-1em}
\paragraph{Communication Compression for Forward Activations and Backward Gradients.}
In slow networks, the communication among 
all machines often becomes the bottleneck~\cite{liu2021distributed}.
To improve the speed of training,
one can conduct \textit{lossy compression} of the data before they are communicated, illustrated as the $C$ module in Figure~\ref{fig:illustration}.
When the model fits into a single machine,
there have been intensive efforts on compressing 
the model gradient~\cite{alistarh2016qsgd,zhang2017zipml,bernstein2018signsgd,wen2017terngrad}. However, when it comes to 
pipeline model parallelism, 
such compression techniques
are less studied. In this paper, we focus on designing efficient communication compression 
algorithms to compress both forward activations
and backward gradients. 
As we will show later, both can be compressed 
significantly with \algname without hurting the model quality.
We also show that it is possible to combine \algname with 
state-of-the-art gradient compression techniques to enable the end-to-end compression
scheme illustrated in Figure~\ref{fig:illustration}.

\vspace{-1em}
\paragraph{Problem Formulation.} In this paper, we 
focus on the following technical problem. 
Note that, for the simplicity of notations, we present here the case where the model 
is partitioned 
onto $K = 2$ machines. 
\algname works for cases with $K > 2$: (1)
in experiments, we consider $K=8$, i.e., a single 
model is partitioned onto 8 machines;
(2) in the supplementary material we provide the theoretical analysis for $K > 2$.

Given a distribution of samples $\cD$, we
consider the following optimization task: 
\begin{align}\label{eqnTask}
\min_{x\in\mathbb{R}^d} \hspace{5mm} f(x) := \mathbb{E}_{\xi \sim \cD} F ( b ( a(\xi, x^{(a)}), x^{(b)}) ),
\end{align}
where 
$F$ is a loss function, 
$a(-)$ and $b(-)$ correspond to two 
sets of \textit{layers} of the model ---
$a(-)$ has model $x^{(a)}$
and $b(-)$ has model $x^{(b)}$.
In Figure~\ref{fig:illustration},
Machine 1 would hold 
$x^{(a)}$ and Machine 2 would hold
$x^{(b)}$. In the following, we 
call the machine that holds 
$x^{(a)}$ Machine $a$ and the 
machine that holds 
$x^{(b)}$ Machine $b$ .
In the standard backpropagation algorithm, the communication between 
these two machines are as follows:
\begin{itemize}
\item Given a data sample $\xi$,
Machine $a$ sends to Machine $b$ the forward activation: $a(\xi, x^{(a)})$ 
\item Machine $b$ sends 
to Machine $a$ the backward gradient on 
$a(\xi, x^{(a)})$.
\end{itemize}

% \vspace{-1em}
\paragraph{Difficulties in Direct Quantization.}

A natural way at compressing forward activations is to send,
instead of $a(\xi, x^{(a)})$,
 a 
 quantized version $m(\xi, x^{(a)}) = Q(a(\xi, x^{(a)}))$.
 This is the quantization scheme that
 state-of-the-art methods such as 
 AC-GC~\cite{evans2021ac} and TinyScript~\cite{fu2020don}
 use. Both AC-GC~\cite{evans2021ac} and TinyScript~\cite{fu2020don} 
assume that gradient is unbiased
when $m(\xi, x^{(a)})$
is an unbiased estimator of $a(\xi, x^{(a)})$.
This enables their convergence rates of $\mathcal{O}(1/\sqrt{T})$. 
However, because of the non-linearity of 
$F$ and $b$ in a deep learning model
with non-linear activation functions, 
an unbiased $m(\xi, x^{(a)})$
\textit{will} lead to biases 
on the gradient. %For example, even in a simple setting where $a(x) = x^2$ and $b(x) = x$, we have that the gradient is unbiased only if $\E [x (Q(x^2) - x^2)] = 0.$
In Appendix, we will provide an example showing that such a gradient bias will hurt SGD convergence 
even 
for a very simple optimization problem.
On the theory side, 
previous efforts on understanding gradient bias~\cite{biasedSGD} have also shown that 
even bounded bias on gradient can impact 
the converges of SGD. 
As we will show later, empirically, this
bias can indeed lead to suboptimal 
models under aggressive compression. 
% \vspace{-1em}
\paragraph{Notation.} Throughout the paper we use the following definitions:
\begin{itemize}\setlength\itemsep{0em}
    \item $f^*$ is the optimal value of $f$.
    \item $N$ is the number of samples.
    \item $x_{t} = (x_{t}^{(a)}, x_{t}^{(b)})$ is the full model at iteration $t$.
    \item $\nabla f(\cdot)$ is the gradient of function $f$.
    \item $g_{\xi_t}(x_t) = \nabla F(\xi_t; x_t)$ is the stochastic gradient.
    \item $Q(\cdot)$ is the quantization function used to compress activations.
    \item $m(\cdot)$ is the message exchanged between $a$ and $b$ in the feed forward path. 
    \item $\|\cdot \|$ denotes the $L_2$-norm. 
\end{itemize}

\begin{algorithm}[t!]
   \caption{\algname Algorithm}
   \label{alg:SHwT}
\begin{algorithmic}[1]
   \STATE {\bfseries Initialize:} $x_0$, learning rate $\gamma$, sub-network $a(-)$ weights $x^{(a)}$, sub-network $b(-)$ weights $x^{(b)}$, quantization function $Q$, array of previous messages $m$ initialized to 0 \\
    \FOR{t = 1, \ldots, T}
    \STATE 
        Randomly sample $\xi_t$ 
        \IF{$\xi_t$ {\bfseries not} seen before}
             \STATE Set $m(\xi_t) = a(\xi_t, x_{t}^{(a)})$
        \ELSE
            \STATE Update $m(\xi_t) \leftarrow m(\xi_t) + Q\big( a(\xi_t, x_{t}^{(a)}) - m(\xi_t)\big)$
        \ENDIF
        \STATE // Machine $a$ sends $Q\big( a(\xi_t, x_{t}^{(a)}) - m(\xi_t)\big)$
        to Machine $b$, which knows $m(\xi_t)$ through a local version of $m$ \\ 
        \STATE Update $x_{t+1}^{(b)} \leftarrow x_{t}^{(b)} - \gamma \cdot \nabla_{x^{(b)}} (f\circ b)\vert_m$ \\
        \STATE // Machine $b$ sends $Q(\nabla_{a} (f\circ b)\vert_m)$ to Machine $a$\\ 
       \STATE  Update $x_{t+1}^{(a)} \leftarrow x_{t}^{(a)} - \gamma \cdot Q(\nabla_{a} (f\circ b)\vert_m) \cdot \nabla_{x^{(a)}} a$ \\
    \ENDFOR
    \STATE {\bfseries Output:} $x = (x_{T}^{(a)}, x_{T}^{(b)})$\\    
\end{algorithmic}
\end{algorithm}
% \vspace{-1em}

\section{\algname: Theoretical Analysis and System Implementations}

In this section we present the \algname, with the goal to  mitigate the above mentioned difficulties that appear in direct quantization of the activation functions. 

\subsection{\algname Algorithm}

% \begin{quote}
% \textit{
% the more training stabilizes $\rightarrow$
% the smaller the changes of the model 
% across epochs
% $\rightarrow$
% the smaller the changes of activations
% for the same training example 
% across epochs
% $\rightarrow$
% the smaller the compression error 
% using the same \#bits
% $\rightarrow$
% training stabilizes more.}
% \end{quote}

% Formally, our algorithm corresponds to the following dynamics. Each sample $\xi$ is fed forward as $\xi \mapsto a(\xi, x^{(a)}) \mapsto m(a(\xi, x^{(a)}) \mapsto b(m(a(\xi, x^{(a)})), x^{(b)}) $, where $x^{(a)}$ and $x^{(b)}$ denote the corresponding weights of $a$ and $b$. Our optimization task becomes the following:
% \begin{equation}\label{eqnMainTask}
% \min_{x^{(a)}, x^{(b)}} \hspace{5mm} f(x) = \mathbb{E}_{\xi \sim \cD} F  \left( b \left( m (a(\xi; x^{(a)}), x^{(b)}) \right) \right),
% \end{equation}
% where $F$ is the loss function. 

Algorithm~\ref{alg:SHwT} illustrates the \algname algorithm. The idea behind Algorithm~\ref{alg:SHwT} is simple --- \textit{instead of quantizing the activations directly, quantize the changes of activations
for the same training example across epochs}. 
As illustrated in Algorithm~\ref{alg:SHwT},
for iteration $t$ and the data sample $\xi_t$, 
\textit{if} it is the first time that
$\xi_t$ is sampled, 
Machine $a$ communicates the full
precision activations without any compression:
$m(\xi_t) = a(\xi_t, x_t^{(a)})$ (Lines 4-5).
Both machines will save $m(\xi_t)$
in a local buffer.
If $\xi_t$ has been
sampled in previous iterations,
Machine $a$ communicates a quantized 
version:
\[
Q(a(\xi_t, x_t^{(a)}) - m(\xi_t)),
\]
where $m(\xi_t)$ was the previous 
message, stored in the local buffer.
Both machines
then update this local buffer:
% \vspace{-0.3em}
\[
m(\xi_t) \leftarrow m(\xi_t) + Q(a(\xi_t, x_t^{(a)}) - m(\xi_t)).
\]
Machine $b$ then use 
$m(\xi_t)$ as the forward activations,
compute backward gradients,
and communicate a quantized 
version of the backward gradient to 
Machine $a$ (Line 11). We use
\[
\delta_\xi = a(\xi_t, x_t^{(a)}) - m(\xi_t)
\]
to denote the \textit{message error} in
sending the activations.

{\bf Update Rules.} 
The above algorithm corresponds to the 
following update rules,
at iteration $t$ with sample $\xi_t$:
% \begin{align*}
% 	x_{t+1}^{(a)}  &= x_{t}^{(a)} - \gamma \cdot \nabla_x (f\circ b)(x,y)_{x=m(\xi, x_{t}^{(a)}); y = x_{t}^{(b)}} \cdot \nabla_x a(\xi, x)_{x=x_{t}^{(a)}}, \\
% 	x_{t+1}^{(b)} &= x_{t}^{(b)} - \gamma \cdot \nabla_y (f \circ b) (x,y)_{x=m (\xi, x_{t}^{(a)}); y = x_{t}^{(b)}},
% \end{align*}
% \vspace{-0.3em}
\begin{align*}
	x_{t+1}^{(a)}  &= x_{t}^{(a)} - \gamma \cdot Q(\nabla_{a} (f\circ b)\vert_{(m(\xi, x_{t}^{(a)}), x_{t}^{(b)})}) \cdot \nabla_{x^{(a)}} a\vert_{x_{t}^{(a)}}, \\
	x_{t+1}^{(b)} &= x_{t}^{(b)} - \gamma \cdot \nabla_{x^{(b)}} (f \circ b) \vert_{(m (\xi, x_{t}^{(a)}), x_{t}^{(b)})},
\end{align*}
where $\gamma$ is the learning rate,
$\nabla_{x^{(b)}} (f \circ b) \vert_{(m (\xi, x_{t}^{(a)}), x_{t}^{(b)})}$
is the gradient on $x^{(b)}$
using the quantized forward
activations $(m (\xi, x_{t}^{(a)})$,
and 
$Q(\nabla_{a} (f\circ b)\vert_{(m(\xi, x_{t}^{(a)}), x_{t}^{(b)})})$ is the quantized 
backward gradient.

Setting $x_{t} = (x_{t}^{(a)}, x_{t}^{(b)})$,
we can rephrase the update rule as
\[
x_{t+1} = x_{t} - \gamma \cdot \left( g_{\xi} (x_t) + \Delta_{\xi} (x_t) \right),
\]
where 
$g_{\xi} (x_t)$ is the stochastic gradient and 
$\Delta_{\xi} (x_t)$ is the \textit{gradient error} 
introduced by communication compression.
We have 
$\Delta_{\xi} = (\Delta_{\xi}^{(a)} + \Delta_{\xi}^{(Q)}, \Delta_{\xi}^{(b)})$ given by:
\begin{align*}
    \Delta_{\xi}^{(Q)} (x_t) &= Q(\nabla_{a} (f\circ b)\vert_{(m(\xi, x_{t}^{(a)}), x_{t}^{(b)})}) \cdot \nabla_{x^{(a)}} a\vert_{x_{t}^{(a)}} - \nabla_{a} (f\circ b)\vert_{(m(\xi, x_{t}^{(a)}), x_{t}^{(b)})} \cdot \nabla_{x^{(a)}} a\vert_{x_{t}^{(a)}}, \\
    \Delta_{\xi}^{(a)} (x_t) &= \nabla_{a} (f\circ b)\vert_{(m(\xi, x_{t}^{(a)}), x_{t}^{(b)})} \cdot \nabla_{x^{(a)}} a\vert_{x_{t}^{(a)}} - \nabla_{a} (f\circ b \circ a)\vert_{(x_{t}^{(a)}, x_{t}^{(b)})} \\
    \Delta_{\xi}^{(b)} (x_t) &= \nabla_{x^{(b)}} (f \circ b)\vert_{(m(\xi, x_{t}^{(a)}), x_{t}^{(b)})} - \nabla_{x^{(b)}} (f \circ b)\vert_{(a(\xi, x_{t}^{(a)}), x_{t}^{(b)})},
\end{align*}
where $ \Delta_{\xi}^{(Q)} (x_t)$ is the error introduced by the gradient quantization in the backpropagation part, whilst  $ \Delta_{\xi}^{(a)} (x_t)$ and $ \Delta_{\xi}^{(b)} (x_t)$ are the errors that the gradients of $a$ and $b$, respectively, inherit from the bias introduced in the forward pass.
% \begin{itemize}
% \item  $ \Delta_{\xi}^{(Q)} (x_t)$: error introduced by the gradient quantization in back propagation 
% \item  $ \Delta_{\xi}^{(a)} (x_t)$: error that the gradient on $a$ inherits from the forward pass 
% \item  $ \Delta_{\xi}^{(b)} (x_t)$: error that the gradient on $b$ inherits from the forward pass 
% \end{itemize}

\subsection{Theoretical Analysis}

We now prove the main theorem which states that, under some standard assumptions that are often used in the literature~\cite{bottou2018optimization,liu2021distributed},
the convergence rate of \algname algorithm is $O(1/\sqrt{T})$ for non-convex objectives, same as 
vanilla SGD.

\paragraph{Assumptions.} We make several assumptions on the networks and the quantization. It is important to note is that we put no restrictions on either the message error $\delta_\xi$, nor the gradient error $\Delta_\xi$.
% \vspace{-1em}
\begin{itemize}[leftmargin=*]
    \item \textbf{(A1: Lipschitz assumptions)} We assume that $\nabla f$, $\nabla (f\circ b)$ and $a$ are $L_{f}$, $L_{f\circ b}$-, and $\ell_a$-Lipschitz, respectively, recalling that a function $g$ is $L_g$-Lipschitz if 
    \[
    \| g(x) -  g(y) \| \leq L_g \| x-y\|, \hspace{5mm} \forall x, \forall y. 
    \]
    Furthermore, we assume that $a$ and $f\circ b$ have gradients bounded by $C_a$ and $C_{f\circ b}$, respectively, i.e. $\| \nabla a(x) \| \leq C_a$, and $\| \nabla (f\circ b) (x) \| \leq C_{f\circ b}$.
    \item \textbf{(A2: SGD assumptions)} We assume that the stochastic gradient $g_\xi$ is unbiased, i.e. $\E_\xi [g_\xi (x)] = \nabla f(x)$, for all $x$, and with bounded variance, i.e. $\E_{\xi} \| g_\xi (x) - \nabla f(x) \|^2 \leq \sigma^2$, for all $x$.
\end{itemize}

\begin{thm}\label{thmMainThm}
Suppose that Assumptions A1, A2 hold,
and consider 
an unbiased quantization function
$Q(x)$ which satisfies that there exists $ c_Q < \sqrt{1/2}$ such that $\E \| x-Q(x) \| \leq c_Q \|x\|$, for all $x$.\footnote{Even for a very simple quantization function $Q(x) = \|x\| \cdot \lceil x/||x|| \rfloor$, where $\lceil \cdot \rfloor$ denotes rounding to the closest $k/2^b$, stochastically, through a simple bound $c_Q = \sqrt{d}/2^b$, 6 bits suffice in a low-dimensional ($\sim10^3$), 11 bits in a high-dimensional scenario ($\sim10^6$), and 16 bits in a super-high-dimensional scenario ($\sim10^9$).
In practice, as we show in 
the experiments, we observe that 
for 2-4 bits are often enough
for fine-tuning GPT-2 style model.
This leaves interesting direction
for future exploration as we expect
a careful analysis of 
sparsity together with 
more advanced 
quantization functions can make this
condition much weaker.
}
Let $\gamma = \frac{1}{3(3L_f+C)\sqrt{T}}$ be the learning rate,
where
\[
C = \frac{4c_{Q} \ell_{a} (1+C_a) L_{f\circ b} N}{\sqrt{1-2c_{Q}^2}}.
\]
Then after performing $T$ updates one has
\begin{equation}\label{eqnMainResult}
    \frac{1}{T} \sum_{t\in [T]} \E \| \nabla f(x_t)\|^2 \lesssim  \frac{(C+L_f)(f(x_1) - f^*)}{\sqrt{T}} + \frac{\sigma^2 + (c_{Q}C_aC_{f\circ b})^2}{\sqrt{T}}.
\end{equation}
\end{thm}

We present the full proof of Theorem~\ref{thmMainThm} in Appendix~\ref{appMainThm}, whereas here we explain the main intuition. The usual starting point in examining convergence rates is to use the fact that $f$ has $L_f$-Lipschitz gradient. It is well known that this implies
\[
\gamma \inn{\nabla f(x_t)}{g_{\xi_t}(x_t)} + f(x_{t+1}) - f(x_t) \leq - \inn{\nabla f(x_t)}{\Delta_{\xi_t}(x_t)} + \frac{\gamma^2L_f}{2} \| g_{\xi_t}(x_t) + \Delta_{\xi_t}(x_t) \|^2.
\]
After taking the expectation over all $\xi_t$ and summing over all $t=1,\ldots,T$, we easily see that the key quantity to bound is $\sum_{t=1}^{T} \E \|\tilde\Delta_{\xi_t}(x_t)\|^2$, where $\tilde\Delta_{\xi_t}(x_t) = (\Delta_{\xi_t}^{(a)}(x_t),\Delta_{\xi_t}^{(b)}(x_t))$. On the other hand, the main object that we can control is the message error, $\delta_\xi$. Therefore, we first prove an auxiliary result which shows that $\|\tilde \Delta_{\xi_t}(x_t) \| \leq (1+C_a) \ell_a \| \delta_{\xi_t}(x_t)\|$, for all $t$. The key arguments for bounding $\delta_{\xi_t}$ closely follow the self-improving loop described in the introduction, and can be summarized as follows. Since we are compressing the information in such a way that we compare with all the accumulated differences, this allows us to unwrap the changes which appeared since the last time that we observed the current sample, in an iterative way. However, since these are gradient updates, they are bounded by the learning rate --- as long as we have a  quantization method that keeps enough information about the signal, we can recursively build enough saving throughout the process. In particular, the more stability we have in the process, the smaller the changes of the model and the compression error gets, further strengthening the stability.

\paragraph{Tightness.} The bound is tight with respect to quantization --- setting $c_Q = 0$ (implying $C=0$), i.e. quantization does not incur any loss, recovers the original original SGD convergence (cf.~\cite{bottou2018optimization,liu2021distributed}).

{
% \color{blue}
\paragraph{Regularization and other optimizers.} Assuming A1 and A2 for $f$, and under further assumptions on $b$ and $\nabla_{x^{(b)}} b$, one can prove that the $L_2$-regularized loss $\tilde{f}(x) = f(x) + \frac{\lambda}{2}\|x\|^2$, which results in weight decay, satisfies Assumptions A1 and A2 with slightly weaker constants. We note that a theoretical analysis for other regularization methods or optimizers such as Adam~\cite{adam2015}, could be an independent study and represent an interesting line of future research.}

\subsection{System Implementations and Optimizations.}

\textbf{Additional storage and communication.} \algname requires us to store, for each 
data example $\xi$, the quantized
activation $m(\xi)$ in a local buffer
in memory or SSD. For example, in GPT2-XL training, a simple calculation shows that we need an approximately extra 1TB storage. When using data parallelism, it reduces to 1TB / $\# $ parallel degree, but also incurs communication overhead if data is shuffled in every epoch. In addition, when the example $\xi$ is 
sampled again,  $m(\xi)$
needs to be (1) loaded from 
this local buffer to the GPU, and 
(2) updated when a new 
value for $m(\xi)$ is ready.

\textbf{Optimization.} It is easy to implement and optimize
the system such that this additional 
loading and updating step
do not incur significant overhead
on the end-to-end runtime.
This is because of the fact that the
GPU computation time for a forward pass is usually much longer than the data transfer time to load the activations --- for GPT2-XL with 
1.5 billion model parameters,
a single forward pass on 6
layers require 44 ms,
whereas loading $m(\xi)$ need 
0.2 ms from memory and
12 ms from SSD.
One can simply pre-fetch 
$m(\xi)$ right before the
forward pass, and hide it within the
forward pass of other data examples.
Similarly, updating $m(\xi)$ can also be hidden in the backward computation. It is also simple to reduce the communication overhead by shuffling data only once or less frequently.

% \textcolor{red}{Additional storage for $m(\xi)$
% XXX\\
% XXX\\
% XXX\\
% GPT-2: one example -- 1.6M number for $m(\xi)$ --- 6.4MB -- 1TB -- 156,250 sentences -- x data parallel degree -- msot of the fine tuning task is fine
% XXX\\
% XXX\\
% XXX\\
% XXX}

%Note that \algname does require 
%additional storage in DRAM and SSD to store 
%$m(\xi)$. 

%\footnote{A detail micro-benchmark about this claim is in Appendix}. 

\section{Evaluation}

We demonstrate that \algname can 
significantly speed up fine-tuning large 
language models in slow networks.
Specifically, we show: (1) 
on four standard benchmark tasks,
\algname can tolerate aggressive 
quantization on the activations (2-4 bits)
and backward gradients
(4-8 bits), without hurting convergence 
and final loss, whereas direct quantization 
converges to a worse loss or even diverge,
(2)
in slow networks, 
\algname provide an end-to-end speed-up 
up to $4.3\times$,
and (3) 
\algname can be combined with state-of-the-art 
gradient compression methods and achieve 
an end-to-end speed-up of 
up to $4.9\times$.

\subsection{Experimental Setup}
\label{exp:setup}

\textbf{Datasets and Benchmarks.} We
consider both \textit{sequence classification} and \textit{language modeling} tasks
with state-of-the-art foundation models.
For sequence classification,
we fine-tune a 1.5B parameter DeBERTa\footnote{we use the v2-xxlarge checkpoint: \url{https://huggingface.co/microsoft/deberta-v2-xxlarge}.} on two datasets: QNLI and CoLA. 
For language modeling, we 
fine-tune the GPT2 model with 1.5B parameters\footnote{we use the extra large checkpoint: \url{https://huggingface.co/gpt2-xl}.} on two datasets: WikiText2 and arXiv abstracts.
All datasets are publicly available and do not contain sensitive or offensive content. 
Detailed setup can be found in Appendix \ref{apd:data}.

\textbf{Distributed Cluster.} 
% We implement GPipe to partition the model into different stages.
We conduct our experiments on AWS
with 8-32 \texttt{p3.2xlarge} instances, each 
containing a V100 GPU. For a single 
pipeline, we partition a model 
onto 8 machines.
When combined with data parallelism, we 
use 32 instances --- data parallel degree is 4 and pipeline parallel degree is 8. 
By default, instances are interconnected with 10Gbps bandwidth. We simulate slow networks
by controlling the communication bandwidth between instances using Linux traffic control.

\textbf{Baselines.}
We compare with several strong baselines:
\begin{enumerate}
\item \texttt{FP32}: in which all communications are in 32 bit floating point without any compression.
\item \texttt{DirectQ}~\cite{evans2021ac,fu2020don}: in which activations and backward gradients are 
directly quantized.
\end{enumerate}
We use a simple, uniform quantization scheme, which first normalizes a given vector into $[-1, 1]$ and quantize each number into a $b$-bit integers
by uniforming partitioning the range $[-1, 1]$
into $2^b$ intervals~\cite{chakrabarti2019backprop}.
Additional details of the configuration can be found in Appendix \ref{apd:setup}.

\textbf{Hyperparameter Tuning.} We conduct careful tuning for all methods on all 
datasets. 
We perform grid search to choose learning rate from $\{$2.5e-6, 3e-6, 5e-6, 1e-5$\}$ 
and macro-batch size from $\{$32, 64, 96$\}$ for best model performance. 
{%\color{blue}
We train all models using the Adam optimizer with weight decay.}

\subsection{Results}

\paragraph*{Convergence.}

\begin{figure}
    \centering
    \begin{subfigure}[b]{0.245\linewidth}
        \centering
        \includegraphics[width=\linewidth]{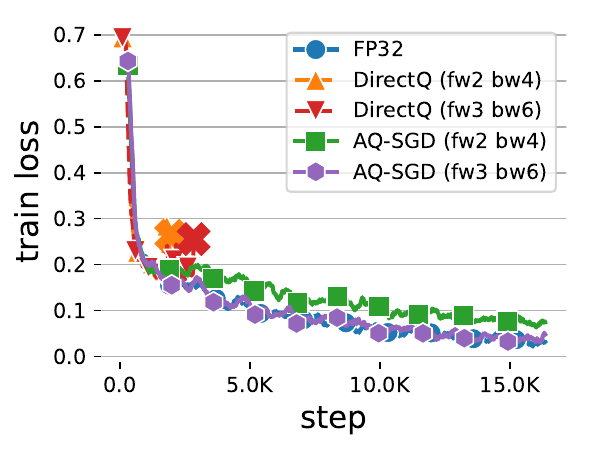}
        % \vspace{-5mm}
        \caption{QNLI, DeBERTa-1.5B}
        \label{fig:exp_qnli}
    \end{subfigure}
    \hfill
    \begin{subfigure}[b]{0.245\linewidth}
        \centering
        \includegraphics[width=\linewidth]{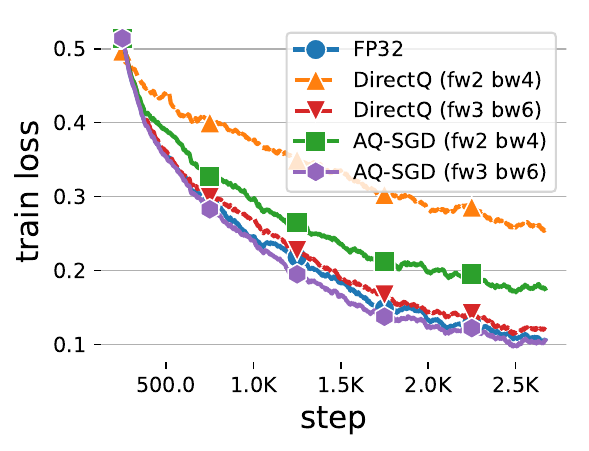}
        % \vspace{-5mm}
        \caption{CoLA, DeBERTa-1.5B}
        \label{fig:exp_cola}
    \end{subfigure}
    \hfill
    \begin{subfigure}[b]{0.245\linewidth}
        \centering
        \includegraphics[width=\linewidth]{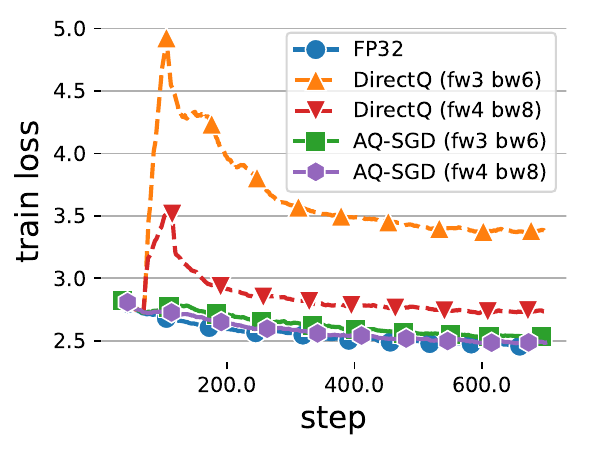}
        % \vspace{-5mm}
        \caption{WikiText2, GPT2-1.5B}
        \label{fig:exp_wiki}
    \end{subfigure}
    \hfill
    \begin{subfigure}[b]{0.245\linewidth}
        \centering
        \includegraphics[width=\linewidth]{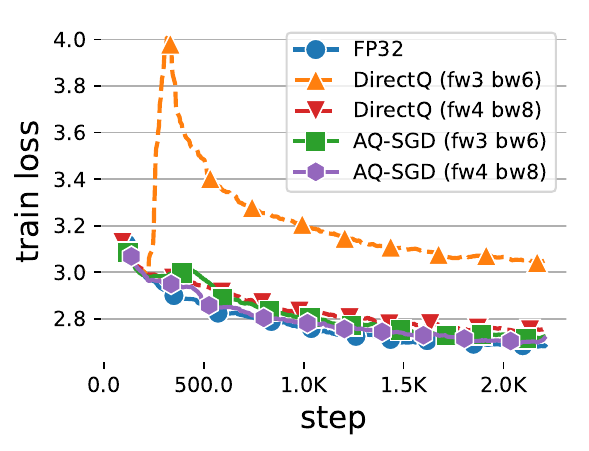}
        % \vspace{-5mm}
        \caption{arXiv, GPT2-1.5B}
        \label{fig:exp_arxiv}
    \end{subfigure}
       \caption{Convergence (loss vs. \# steps) of different approaches. $\boldsymbol{\times}$ represents divergence.}
       \label{fig:exp_convs}
\end{figure}

% We first compares the training loss of different methods as a function of the number of iterations.
We first compare the convergence behavior 
of different methods. For all compression methods, we try various settings: \texttt{fw$x$ bw$y$} means that we use $x$ bits for 
forward activation and $y$ bits for backward gradients.
Figure \ref{fig:exp_convs} shows the convergence behavior for the sequence classification and language modeling tasks.
\texttt{FP32} converges fast since it does not introduce any compression errors. \texttt{DirectQ}, under
aggressive quantization, 
can converge to a significantly worse 
model, or even diverge. 
This is not surprising, given the 
biases on model gradients that direct quantization introduced.
On the other hand, \algname converges almost as fast as \texttt{FP32} in terms of number of training steps. 
% it achieves the best convergence performance on all datasets.
%However, it requires transferring a large volume of data for each mini-batch, which can significantly slow down training especially when network bandwidth is limited.
%\texttt{Uniform} reduces the amount of communicated data but it also hurts the convergence performance and sometimes fails to converge.
%This shows that directly applying quantization methods to activations will inevitably bring cause high quantization variance and thus cannot achieves acceptable performance.
%In comparison, \algname only has to compress and communicate the difference between current and previously seen activations, which is smaller in magnitude compared with activations themselves.
%And it achieves sound convergence performance with an aggressive compression setting (2-4 bits).
%Also, an interesting phenomenon is that due to compression error, the loss of \algname rises slightly when warm-up is complete, but it converges quickly with more training epochs.
%This is consistent with our theory that the updates to the activations become smaller, and therefore the compression error also becomes smaller.

\begin{figure}
    \centering
    % \vspace{-4mm}
    \begin{subfigure}[b]{0.245\linewidth}
        \centering
        \includegraphics[width=\linewidth]{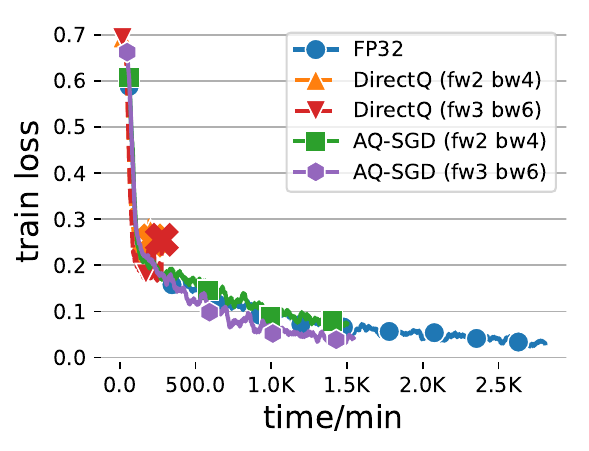}
        % \vspace{-5mm}
        \caption{QNLI, 500Mbps}
        \label{fig:exp_e2e_qnli_500Mb}
    \end{subfigure}
    \hfill
    \begin{subfigure}[b]{0.245\linewidth}
        \centering
        \includegraphics[width=\linewidth]{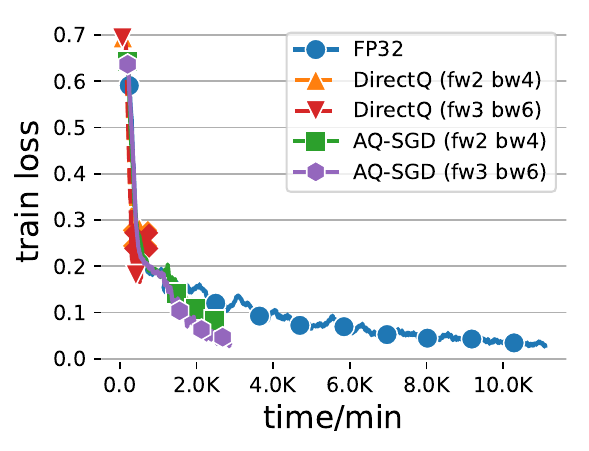}
        % \vspace{-5mm}
        \caption{QNLI, 100Mbps}
        \label{fig:exp_e2e_qnli_100Mb}
    \end{subfigure}
    \hfill
    \begin{subfigure}[b]{0.245\linewidth}
        \centering
        \includegraphics[width=\linewidth]{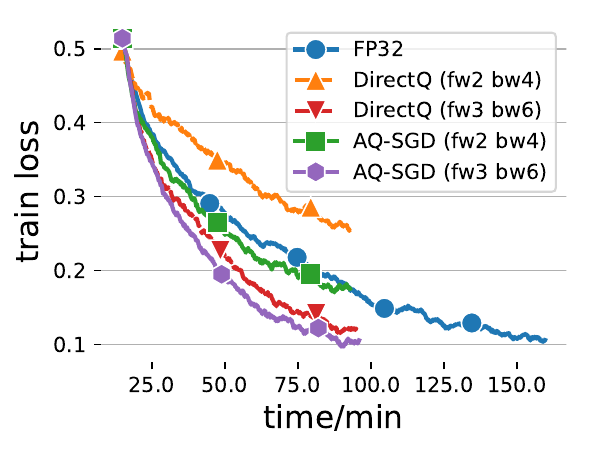}
        % \vspace{-5mm}
        \caption{CoLA, 500Mbps}
        \label{fig:exp_e2e_cola_500Mb}
    \end{subfigure}
    \hfill
    \begin{subfigure}[b]{0.245\linewidth}
        \centering
        \includegraphics[width=\linewidth]{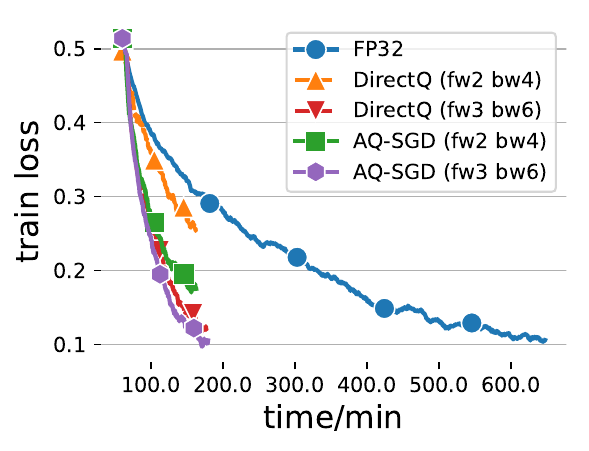}
        % \vspace{-5mm}
        \caption{CoLA, 100Mbps}
        \label{fig:exp_e2e_cola_100Mb}
    \end{subfigure}
    % \vspace{-1mm}
    
    \begin{subfigure}[b]{0.245\linewidth}
        \centering
        \includegraphics[width=\linewidth]{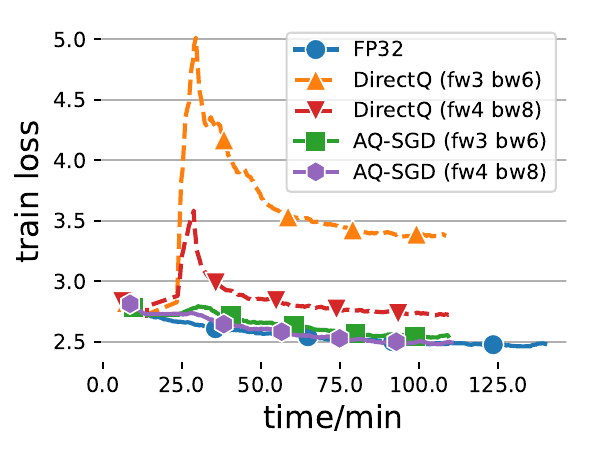}
        % \vspace{-5mm}
        \caption{WikiText2, 500Mbps}
        \label{fig:exp_e2e_wiki_500Mb}
    \end{subfigure}
    \hfill
    \begin{subfigure}[b]{0.245\linewidth}
        \centering
        \includegraphics[width=\linewidth]{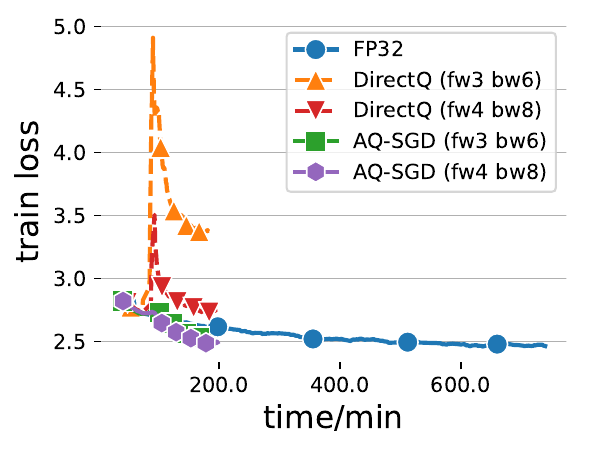}
        % \vspace{-5mm}
        \caption{WikiText2, 100Mbps}
        \label{fig:exp_e2e_wiki_100Mb}
    \end{subfigure}
    \hfill
    \begin{subfigure}[b]{0.245\linewidth}
        \centering
        \includegraphics[width=\linewidth]{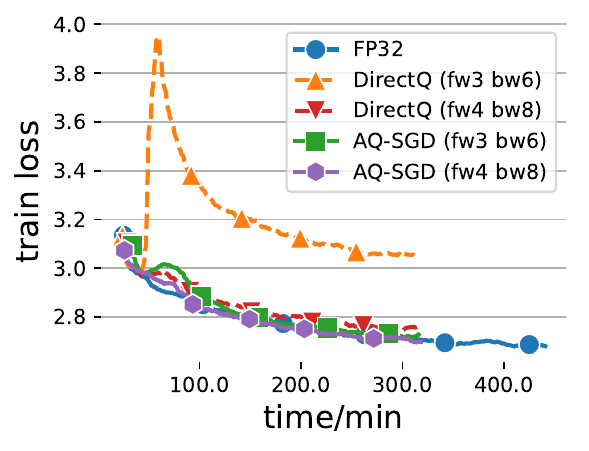}
        % \vspace{-5mm}
        \caption{arXiv, 500Mbps}
        \label{fig:exp_e2e_arxiv_500Mb}
    \end{subfigure}
    \hfill
    \begin{subfigure}[b]{0.245\linewidth}
        \centering
        \includegraphics[width=\linewidth]{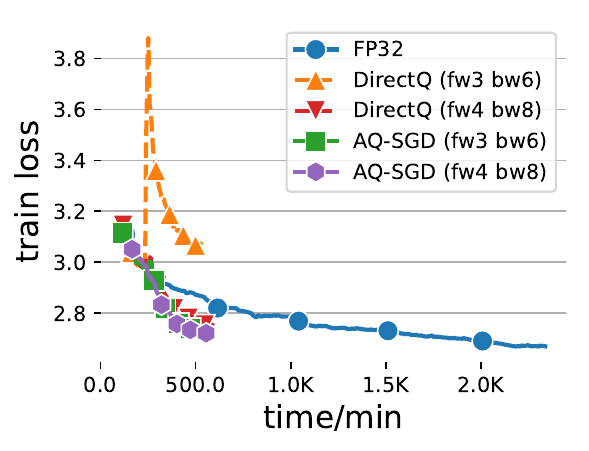}
        % \vspace{-5mm}
        \caption{arXiv, 100Mbps}
        \label{fig:exp_e2e_arxiv_100Mb}
    \end{subfigure}
    % \vspace{-5mm}
    
       \caption{End-to-end training performance over different networks. $\boldsymbol{\times}$ represents divergence.}
       \label{fig:exp_e2e}
\end{figure}

\paragraph*{End-to-End Runtime.}
We show the end-to-end runtime of different methods under slow networks. As illustrated in 
Figure~\ref{fig:exp_e2e},  \algname achieves a 4.3$\times$ end-to-end speed-up comparing with that of \texttt{FP32} (in terms of time to the same loss), illustrating the importance of 
communication compression in slow networks.
Table~\ref{tab:efficiency} shows the 
training throughput 
{
% \color{blue}
and Table~\ref{tab:breakdown} shows the breakdown of our algorithm.
We note that computation and communication can overlap, so the end-to-end time depends on the larger one of the two.
}
Another interesting observation is that when the network is $100\times$ slower (from 10Gbps to 100Mbps), the training is only about $1.18\times$ slower! This is exciting --- if \algname were to be deployed in a in real-world geo-distributed decentralized
networks, the training throughput would be almost as fast as the performance inside a data center!

Moreover, \algname does not introduce 
significant runtime overhead over direct quantization. From Table~\ref{tab:efficiency},
we see that \algname is essentially as efficient as direct quantization compression in throughput.

%which demonstrates the effectiveness of \algname in improving the training efficiency in real-world settings.
%We measure the hardware efficiency under different network conditions inserted by Linux traffic control tools.
%Table \ref{tab:efficiency} compares the throughput under different network configurations. 
%Taking into consideration the warm-up phase, \algname offers 4.7-6.4$\times$ communication volume reduction.
%This enables 5-6$\times$ higher throughput than uncompressed baseline under slow networks.
%Note that \texttt{FP32} is a strong baseline based on our asynchronous implementation of GPipe, where communication and computation can overlap to some extend.
%However, as shown in Figure \ref{fig:exp_e2e}, \algname achieves a 4.3$\times$ end-to-end speed-up in training time compared to \texttt{FP32},
%which demonstrates the effectiveness of \algname in improving the training efficiency in real-world settings.

% We note that in our setup, the throughput of DeBERTa is smaller than that of GPT2 
% mainly because we set a smaller sequence length, which is 256 compared to GPT2's 1024.

% \vspace{-5mm}

\begin{table}[t!]
    \parbox[t]{0.49\linewidth}{
    \centering
    \small
    % \scriptsize
    % \vspace{-4mm}
    \renewcommand{\arraystretch}{0.95}
    \caption{Training Throughput of GPT2-1.5B. Others are similar and shown in Appendix.}
    \setlength\tabcolsep{4 pt}
    \begin{tabular}{@{}rcccccc@{}}
    \toprule
        % ~ &    \multicolumn{3}{c}{GPT2-1.5B} \\  \cmidrule(lr){2-4} 
        \begin{tabular}{@{}c@{}}
            Network \\ Bandwidth
        \end{tabular}
        & \texttt{FP32} & 
        \begin{tabular}{@{}c@{}}
            \texttt{DirectQ} \\ \cmidrule{0-0}
            {\scriptsize fw3 bw6 / fw4 bw8 }
        \end{tabular}
        & \begin{tabular}{@{}c@{}}
            \algname \\ \cmidrule{0-0}
            {\scriptsize fw3 bw6 / fw4 bw8 }
        \end{tabular}   
        \\
    \midrule             
        10 Gbps  & 3.8   &  4.0 \,\, / \,\,  4.1     &  4.0 \,\, / \,\,  4.0   \\
        1 Gbps   & 3.2   &  4.0 \,\, / \,\,  4.0     &  4.0 \,\, / \,\,  3.9   \\
        500 Mbps & 2.7   &  3.9 \,\, / \,\,  3.9     &  3.9 \,\, / \,\,  3.9   \\
        300 Mbps & 1.8   &  3.9 \,\, / \,\,  3.8     &  3.8 \,\, / \,\,  3.8  \\
        100 Mbps & 0.5   &  3.5 \,\, / \,\,  3.0     &  3.4 \,\, / \,\,  3.0  \\
    \bottomrule \\
    \end{tabular}
    
    \label{tab:efficiency}
    % \vspace{-5mm}
    }
    \hfill
    \parbox[t]{0.49\linewidth}{
    % \color{blue}
    % \captionsetup{labelfont={color=blue}}
    \centering
    \small
    % \scriptsize
    % \vspace{-4mm}
    \renewcommand{\arraystretch}{0.92}
    \caption{
            % \color{blue} 
            Breakdown of \algname (fw4 bw8) on GPT2-1.5B.
            We show the computation and communication time of each micro batch.}
    \setlength\tabcolsep{5.5 pt}
    \begin{tabular}{@{}rcccccc@{}}
    \toprule
        \multirow{2}[2]{*}{
        \begin{tabular}{@{}c@{}}
            Network \\ Bandwidth
        \end{tabular} 
        }
                 & \multicolumn{2}{c}{Forward pass} & \multicolumn{2}{c}{Backward pass} \\ \cmidrule(lr){2-3} \cmidrule(lr){4-5}  
                 & comp. & comm. & comp. & comm.  \\
    \midrule             
        % 1 Gbps   & 45 ms  & \, \,8 ms   & 135 ms  & \, \,15 ms   \\
        500 Mbps & 45 ms  & 13 ms       & 135 ms  & \, \,25 ms   \\
        300 Mbps & 45 ms  & 21 ms       & 135 ms  & \, \,42 ms  \\
        200 Mbps & 45 ms  & 31 ms       & 135 ms  & \, \,63 ms  \\
        100 Mbps & 45 ms  & 63 ms       & 135 ms  & 125 ms  \\
    \bottomrule \\
    \end{tabular}
    
    \label{tab:breakdown}
    }
    
\end{table}

\subsection{End-to-end Communication Compression: \algname + QuantizedAdam} 
\label{exp:dp_pp}

\algname can also be combined with 
existing methods on gradient compression.
This allows us to compress \textit{all}
communications during training.
We combine \algname with QuantizedAdam~\cite{tang20211}, an error compensation-based gradient compression algorithm for data parallel training.  

%In this subsection, we show that \algname works well with gradient compression when both data parallelism and pipeline parallelism are used.
%We focus on fine-tuning GPT2-1.5B with 32 AWS p3.2xlarge instances,
%where the data parallel group size is 4 and the pipeline group size is 8. 
%We set the global mini-batch size to 96, i.e.~24 for the local mini-batch size.
%We perform 4-bit quantization to reduce communication overhead in data parallelism with error compensation \cite{seide20141}.
%Following \cite{tang20211}, we freeze the Adam variance after several iterations of uncompressed training, and we compress and exchange the momentums of parameters instead of gradients.

\begin{figure}
    \centering
    \begin{subfigure}[b]{0.27\linewidth}
        \centering
        \includegraphics[width=\linewidth]{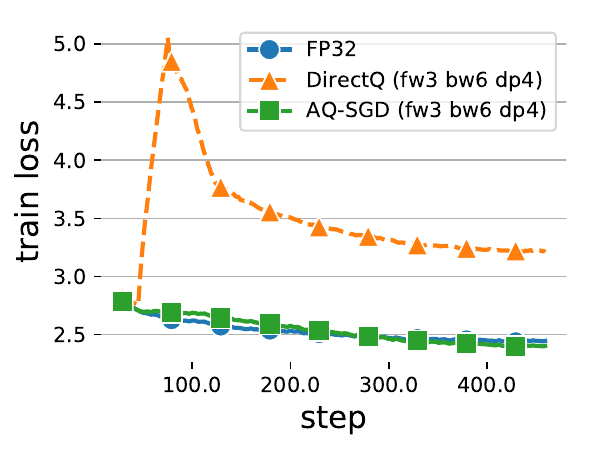}
        % \vspace{-6mm}
        \caption{WikiText2, GPT2-1.5B}
    \end{subfigure}
    \hfill
    \begin{subfigure}[b]{0.27\linewidth}
        \centering
        \includegraphics[width=\linewidth]{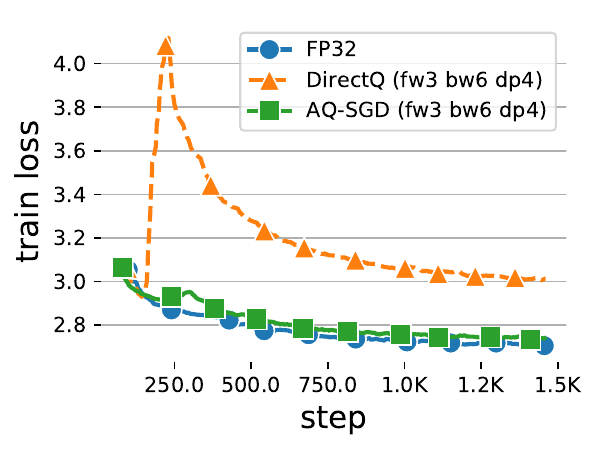}
        % \vspace{-6mm}
        \caption{arXiv, GPT2-1.5B}
    \end{subfigure}
    \hfill
    \begin{subfigure}[b]{0.41\linewidth}
        \centering
        \includegraphics[width=\linewidth]{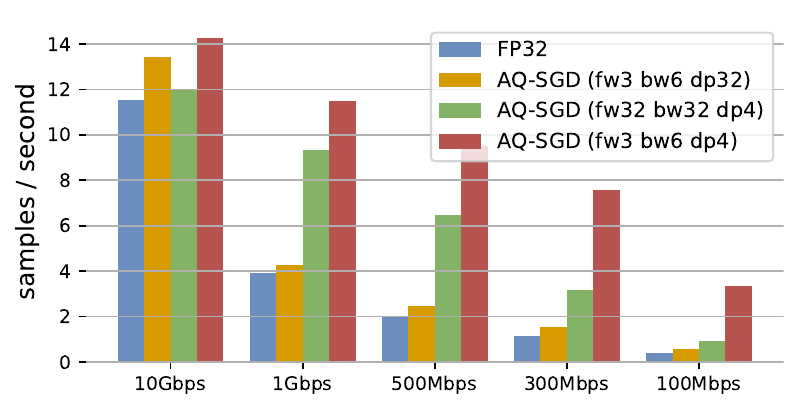}
        % \vspace{-6mm}
        \caption{Training Throughput}
    \end{subfigure}
    % \vspace{-0.2em}
    \caption{Convergence and Throughput of \algname combined with gradient compression.}
    % \vspace{-0.5em}
    \label{fig:dp_pp}
\end{figure}

We quantize the forward activations 
with 3 bits,
the backward gradient with
6 bits,
and model gradient with
4 bits.
Figure~\ref{fig:dp_pp} illustrates the convergence and the training throughput under different network configuration.
We see that \algname converges well when combined with QuantizedAdam (Figure~\ref{fig:dp_pp}(a, b)).
On the other hand, \texttt{DirectQ}, when combined with 
gradient compression, converges to a
much worse model.
In terms of training throughput,
with both activation and gradient compression,
we can achieve up to 8.5$\times$ throughput improvement compared to the no-compression baseline (Figure~\ref{fig:dp_pp}(c)). 
We also see that \textit{both} activation and gradient compression are important in terms of improving end-to-end training throughput ---
as illustrated in Figure~\ref{fig:dp_pp}(c),
disabling any of them will lead to a much 
lower training throughput.

%Thus, for the first time, we provide an end-to-end communication-efficient distributed training solution where all data is communicated with low precision without compromising convergence performance.

\section{Related Work}

\textbf{Distributed training of foundation models.} Modern distributed training of deep neural networks goes beyond data parallelism~\cite{shoeybi2019megatron,huang2019gpipe,yang2021pipemare} due to the advance of the large-scale foundation models \cite{DBLP:journals/corr/abs-2108-07258}, such as BERT~\cite{devlin2018bert}, GPT-3~\cite{brown2020language}, and CLIP\cite{radford2021learning}. Popular systems to support foundation model training include Megatron~\cite{shoeybi2019megatron}, Deepspeed~\cite{rasley2020deepspeed}, and Fairscale~\cite{baines2021fairscale}.
To scale out the training of large-scale models, pipeline parallelism (e.g., Gpipe\cite{huang2019gpipe}, Pipedream\cite{narayanan2019pipedream, narayanan2021memory}) is a popular option, where the model is partitioned into multiple stages, different stages are allocated on different GPUs and the exchange of activations and gradients on activations goes through network communication.

%Split learning \cite{gupta2018distributed,vepakomma2020nopeek,vepakomma2018split} protects data privacy by splitting the model between edge devices and data center, and communicating activation instead of raw data. In these scenarios, the overhead of communicating activation in forward propagation and corresponding gradients of activation in backward propagation can become the bottleneck of distributed learning. 

\textbf{Communication compression of distributed learning.} Communication compression is an effective system relaxation for distributed training, especially in data parallelism~\cite{tang20211,chen2020scalecom,xie2020cser,faghri2020adaptive,li2020acceleration,rothchild2020fetchsgd,safaryan2021stochastic,gorbunov2021marina,safaryan2021smoothness,qian2021error,m2021efficient,wang2021pufferfish}. Popular techniques include quantization~\cite{alistarh2016qsgd,zhang2017zipml,bernstein2018signsgd,wen2017terngrad}, 
sparsification~\cite{wangni2018gradient,alistarh2018convergence,wang2018atomo,wang2017efficient}, sketching~\cite{sketchml,ivkin2019communication} and 
error compensation~\cite{tang2019doublesqueeze}. Recently, TinyScript~\cite{fu2020don} proposes to compress activations and gradients simultaneously.  

\textbf{Sparse Learning for activation compression.} Sparse learning \cite{fu2020don,mocanu2018scalable,raihan2020sparse, chen2021pixelated,dettmers2019sparse,mostafa2019parameter,junjie2019dynamic,hoefler2021sparsity,liu2021sparse} has become increasingly popular for training neural networks, as it can significantly reduce the use of computation and memory while preserving the generalization of such models. In particular, activation compression methods \cite{han2016deep,hubara2017quantized,jain2018gist,chakrabarti2019backprop,chen2021actnn, evans2021ac} are proposed to reduce the memory footprint by adopting lossless \cite{choukse2020buddy,rhu2018compressing,lascorz2019shapeshifter} and lossy \cite{jin2021novel,evans2021ac,evans2020jpeg} compression techniques in the training of various deep neural networks (e.g., CNN\cite{mostafa2019parameter,georgiadis2019accelerating,gudovskiy2018dnn}, GNN\cite{liu2021exact}). These approaches usually compute the activation with \textit{full precision} in forward propagation, adopt the compression method over the activation, and store the compressed version in DRAM for later use in backward propagation. Thus, compression does not introduce any error in forward propagation in contrast to the scenario of communicating compressed activation in the distributed setting. 

\textbf{Delta-based compression.} Delta-based compression \cite{pekhimenko2012base} is a classic solution to various system problems. Recent research has also used the property of spatial proximity within activation in neural network training based on empirical observations \cite{awad2021fpraker, bersatti2020neural,evans2020jpeg}. However, to our knowledge, no attempt has been made to consider the proximity of activation 
through training epochs to enable efficient compression with theoretical guarantee.

\section{Conclusion}
In this paper, we discuss how to adopt communication compression for activations in distributed learning. We proposed \algname, a novel activation compression algorithm for communication-efficient pipeline parallelism training over slow networks. \algname achieves $O(1/\sqrt{T})$  convergence rate for non-convex optimization without making assumptions on gradient unbiasedness. Our empirical study suggests that \algname can achieve up to $4.3\times$ speedup for pipeline parallelism. When combined with gradient compression in data parallelism, the end-to-end speed-up can be up to $4.9\times$.

\section*{Acknowledgments}
\begin{footnotesize}
CZ and the DS3Lab gratefully acknowledge the support from the Swiss State Secretariat for Education, Research and Innovation (SERI) under contract number MB22.00036 (for European Research Council (ERC) Starting Grant TRIDENT 101042665), the Swiss National Science Foundation (Project Number 200021\_184628, and 197485), Innosuisse/SNF BRIDGE Discovery (Project Number 40B2-0\_187132), European Union Horizon 2020 Research and Innovation Programme (DAPHNE, 957407), Botnar Research Centre for Child Health, Swiss Data Science Center, Alibaba, Cisco, eBay, Google Focused Research Awards, Kuaishou Inc., Oracle Labs, Zurich Insurance, and the Department of Computer Science at ETH Zurich. CR gratefully acknowledges the support of NIH under No. U54EB020405 (Mobilize), NSF under Nos. CCF1763315 (Beyond Sparsity), CCF1563078 (Volume to Velocity), and 1937301 (RTML); ARL under No. W911NF-21-2-0251 (Interactive Human-AI Teaming); ONR under No. N000141712266 (Unifying Weak Supervision); ONR N00014-20-1-2480: Understanding and Applying Non-Euclidean Geometry in Machine Learning; N000142012275 (NEPTUNE); NXP, Xilinx, LETI-CEA, Intel, IBM, Microsoft, NEC, Toshiba, TSMC, ARM, Hitachi, BASF, Accenture, Ericsson, Qualcomm, Analog Devices, Google Cloud, Salesforce, Total, the HAI-GCP Cloud Credits for Research program, the Stanford Data Science Initiative (SDSI), and members of the Stanford DAWN project: Facebook, Google, and VMWare. The U.S. Government is authorized to reproduce and distribute reprints for Governmental purposes notwithstanding any copyright notation thereon. Any opinions, findings, and conclusions or recommendations expressed in this material are those of the authors and do not necessarily reflect the views, policies, or endorsements, either expressed or implied, of NIH, ONR, or the U.S.
This work was done during Jue Wang's visit to ETH Z\"urich, 
which was supported by Key Research and Development Program of Zhejiang Province of China (No. 2021C01009) and Fundamental Research Funds for the Central Universities.
The computation required in this work was provided by Together Computer (\url{https://together.xyz/}).
\end{footnotesize}

\bibliographystyle{unsrtnat}
\bibliography{main}

\clearpage
\appendix

\section{Proof of the Main Theorem}\label{appMainThm}

In this section we prove Theorem~\ref{thmMainThm}. The main idea comes from the ``self-enforcing'' dynamics described in the introduction of this work: 
\textit{the more training stabilizes $\rightarrow$
the smaller the changes of the model 
across epochs
$\rightarrow$
the smaller the changes of activations
for the same training example 
across epochs
$\rightarrow$
the smaller the compression error 
using the same \#bits
$\rightarrow$
training stabilizes more.}

We start by providing two auxiliary results. The first one connects the message error and the gradient error. 

\begin{lem}\label{lemAuxOne}
For every sample $\xi$, one has 
\[
\|\Delta_{\xi}^{(Q)} (x)\| \leq c_Q C_a C_{f\circ b},
\]
and 
\[
\|\tilde\Delta_\xi (x)\| \leq (1+C_a) L_{f \circ b} \|\delta_\xi (x)\|,
\]
where $\tilde \Delta_\xi (x) = (\Delta_{\xi}^{(a)}(x), \Delta_{\xi}^{(b)})$. 
\end{lem}
\begin{proofcap}
Note that
\begin{align*}
    \| \Delta_{\xi}^{(Q)} (x_t)\| &= \| \nabla_x a(\xi, x)_{x=x_{t}^{(a)}} \| \cdot \| Q(\nabla_x (f\circ b)(x,x_{t}^{(b)})_{x=m(\xi, x_{t}^{(a)})}) - \nabla_x (f\circ b)(x, x_{t}^{(b)})_{x=m(\xi, x_{t}^{(a)})} \| \\
    &\leq C_a c_Q \| \nabla_x (f\circ b)(x,x_{t}^{(b)})_{x=m(\xi, x_{t}^{(a)})} \| \leq c_Q C_a C_{f\circ b},\\
    \| \Delta_{\xi}^{(a)} (x_t) \| &= \| \nabla_x a(\xi, x)_{x=x_{t}^{(a)}} \| \cdot \| \nabla_x (f\circ b)(x,x_{t}^{(b)})_{x=m(\xi, x_{t}^{(a)})} - \nabla_x (f\circ b)(x, x_{t}^{(b)})_{x=a(\xi, x_{t}^{(a)})} \| \\
    &\leq C_a L_{ f\circ b} \| (m(\xi,x_{t}^{(a)}), x_{t}^{(b)}) - ( a(\xi, x_{t}^{(a)}),x_{t}^{(b)}) \| = C_a L_{ f\circ b} \| \delta_\xi (x_t)\| , \\
    \| \Delta_{\xi}^{(b)}  (x_t) \| &= \| \nabla_y (f \circ b) (m(\xi, x_{t}^{(a)}),y)_{y = x_{t}^{(b)}} - \nabla_y (f \circ b) (a(\xi, x_{t}^{(a)}),y)_{y = x_{t}^{(b)}} \| \\
    &\leq L_{ f\circ b}\| (m(\xi,x_{t}^{(a)}), x_{t}^{(b)}) - ( a(\xi, x_{t}^{(a)}),x_{t}^{(b)}) \| = L_{ f\circ b}\| \delta_\xi (x_t)\|,
\end{align*}
which together with $\| \tilde \Delta_\xi(x_t) \| = \| \Delta_{\xi}^{(a)}(x_t) \| + \| \Delta_{\xi}^{(b)} (x_t) \|$ yields the claim. 
\end{proofcap}

We now prove that the message error can be efficiently bounded by the true gradient.

\begin{lem}\label{lemAuxTwo}
For $C' = \frac{18c_{Q}^2 l_{a}^2 N^2}{1-2c_{Q}^2}$, one has
\[
\frac{1}{T} \sum_{t\in [T]} \E \| \delta_{\xi_t}(x_t) \|^2 \leq C' \gamma^2 \cdot \left( \frac{1}{T} \sum_{t\in [T]} \E \| \nabla f (x_t)\|^2 + \sigma^2 + (c_Q C_a C_{f\circ b})^2\right).
\]
\end{lem}
\begin{proofcap}
    Let $\xi$ be a fixed sample. To simplify the exposition, we abuse the notation slightly by $a(x) = a(\xi,x)$, $m(x) = m(\xi,x)$. Let $T(\xi)$ be the number of realizations of $\xi$ before time $T$. Using the definition of $\delta_\xi$ (recalling that $\delta_\xi (x_{t_1(\xi)}) = 0$, since in the first iteration we send the correct signal), we have
    \begin{align*}
        \sum_{k=1}^{T(\xi)} \| \delta_\xi (x_{t_k(\xi)}) \|^2 &= \sum_{k=2}^{T(\xi)} \|a(x_{t_{k}(\xi)}) - m(x_{t_{k}(\xi)})\|^2 \\
        &= \sum_{k=2}^{T(\xi)}\|a(x_{t_{k}(\xi)}) - m(x_{t_{k-1}(\xi)}) - Q(a(x_{t_{k}(\xi)}) - m (x_{t_{k-1}(\xi)}))\|^2 \\
        \{ \| x- Q(x) \| \leq c_Q \|x\| \} \hspace{10mm} &\leq c_{Q}^2 \sum_{k=2}^{T(\xi)} \| a(x_{t_{k}(\xi)}) - a(x_{t_{k-1}(\xi)}) + \delta_{\xi} (x_{t_{k-1} (\xi)})\|^2 \\
        \{ (\alpha + \beta)^2 \leq 2\alpha^2 + 2\beta^2 \} \hspace{10mm} &\leq 2c_{Q}^2\sum_{k=2}^{T(\xi)} \| a(x_{t_{k}(\xi)}) - a(x_{t_{k-1}(\xi)}) \|^2 + 2c_{Q}^2\sum_{k=2}^{T(\xi)} \|\delta_\xi(x_{t_{k-1}(\xi)})\|^2 \\
        \{ a \text{ is }\ell_a \text{-Lipschitz } \} \hspace{10mm} &\leq 2c_{Q}^2 \ell_{a}^2 \sum_{k=2}^{T(\xi)} \| x_{t_{k}(\xi)} - x_{t_{k-1}(\xi)}\|^2 + 2c_{Q}^2\sum_{k=2}^{T(\xi)} \|\delta_\xi(x_{t_{k-1}(\xi)})\|^2.
    \end{align*}
    Transferring the $\delta_\xi$ part to the LHS, and noting that between every two realizations of $\xi$ at times $t_{k-1}(\xi)$ and $t_{k}(\xi)$, we can follow updates for $t = t_{k-1}(\xi), \ldots, t_{k}(\xi)-1$, we get
	\begin{align*}
	(1-2c_{Q}^2) \sum_{k=1}^{T(\xi)} \| \delta_\xi (x_{t_k(\xi)}) \|^2 &\leq \gamma^2 \cdot (2c_{Q}^2 \ell_{a}^2) \sum_{k=1}^{T(\xi)} \left\| \sum_{t = t_{k-1}(\xi)}^{t_{k}(\xi) -1} \left( g_{\xi_t}(x_t) + \Delta_{\xi_t}(x_t) \right) \right\|^2 \\
	\{ \text{Cauchy-Schwarz} \} \hspace{5mm} &\leq \gamma^2 \cdot (2c_{Q}^2 \ell_{a}^2) \sum_{k=1}^{T(\xi)} (t_{k}(\xi)-t_{k-1}(\xi)) \sum_{t = t_{k-1}(\xi)}^{t_{k}(\xi) -1} \left\|  g_{\xi_t}(x_t) + \Delta_{\xi_t}(x_t) \right\|^2 \\
	&= \gamma^2 \cdot (2c_{Q}^2 \ell_{a}^2) \sum_{t\in[T]} \omega_\xi (t) \cdot \| g_{\xi_t}(x_t) + \Delta_{\xi_t}(x_t) \|^2,
	\end{align*}
	where $\omega \colon [T] \rightarrow \{0, 1, \ldots \}$ is defined by $\omega_\xi(t) = t_{k}(\xi) - t_{k-1}(\xi)$, if $t\in[t_{k-1}(\xi), t_{k}(\xi)-1]$, and $\omega_\xi(t)=0$, if $t>t_{T(\xi)}(\xi)$. We note that for two different samples $\xi$ and $\xi'$, the sums on the LHS are disjoint. Therefore, summing over all samples $\xi$ and taking the expectation over $\xi$ and all $\xi_t$ yields
	\begin{align*}
	(1-2c_{Q}^2) \cdot \frac{1}{T}\sum_{t\in [T]} \E \| \delta_{\xi_t} (x_{t})\|^2 &\leq \gamma^2 \cdot (2c_{Q}^2\ell_{a}^2) \cdot \frac{1}{T}\sum_{t\in[T]} \E \| g_{\xi_t}(x_t) + \Delta_{\xi_t}(x_t)\|^2 \cdot N \cdot \E [\omega_\xi(t)] \\
	&\leq \gamma^2 \cdot (2c_{Q}^2\ell_{a}^2 N^2) \cdot \frac{1}{T}\sum_{t\in[T]} \E \| g_{\xi_t}(x_t) + \Delta_{\xi_t}(x_t)\|^2 ,
	\end{align*}
	since $\E [\omega_\xi (t)] \leq N$. Applying $\|g_{\xi_t}(x_t) + \Delta_{\xi_t}(x_t)\|^2 \leq 3 \|g_{\xi_t}(x_t)\|^2 + 3 \| \Delta_{\xi_t}^{(Q)}(x_t)\|^2  + 3 \| \tilde\Delta_{\xi_t}(x_t)\|^2$, bounded variance $\E\| g_{\xi}(x) - \nabla f (x)\|^2 \leq \sigma^2$, and Lemma~\ref{lemAuxOne}, we get
	\begin{align*}
	&\left( 1-2c_{Q}^2 - \gamma^2 \cdot 6c_{Q}^2\ell_{a}^2N^2 (1+C_a)^2 L_{f\circ b}^{2}  \right) \cdot \frac{1}{T}\sum_{t\in [T]} \E \| \delta_{\xi_t} (x_{t})\|^2 \\ 
	&\hspace{30mm}\leq \gamma^2 \cdot (6c_{Q}^2\ell_{a}^2 N^2) \left( 2\sigma^2 + 2\cdot \frac{1}{T}\sum_{t\in [T]} \E \| \nabla f (x_t) \|^2 + (c_{Q} C_{a} C_{f\circ b})^2\right),
	\end{align*}
	which implies 
	\begin{align*}
	&\left( 1-2c_{Q}^2 - \gamma^2 \cdot \frac{3}{8} \cdot C^2\cdot (1-2c_{Q}^2) \right) \cdot \frac{1}{T}\sum_{t\in [T]} \E \| \delta_{\xi_t} (x_{t})\|^2 \\ 
	&\hspace{30mm}\leq \gamma^2 \cdot (12 c_{Q}^2\ell_{a}^2 N^2) \left( \sigma^2 +  \frac{1}{T}\sum_{t\in [T]} \E \| \nabla f (x_t) \|^2 + (c_{Q} C_{a} C_{f\circ b})^2\right),
	\end{align*}	
	Recalling the definitions of $C$ and $\gamma$, and the fact that $\gamma \cdot C < \frac{1}{3}$, we can simplify the LHS to get
	\begin{align*}
	&\left( 1-2c_{Q}^2 \right) \cdot \frac{1}{T}\sum_{t\in [T]} \E \| \delta_{\xi_t} (x_{t})\|^2 \leq \gamma^2 \cdot (12\cdot \frac{24}{23} \cdot c_{Q}^2\ell_{a}^2 N^2) \left( \sigma^2 + \frac{1}{T}\sum_{t\in [T]} \E \| \nabla f_{\xi_t}(x_t) \|^2 + (c_{Q} C_{a} C_{f\circ b})^2\right),
	\end{align*}
	yielding the claim.
\end{proofcap}

{
% \color{red}
We are ready to prove the main result, with a learning rate of 
\[
\gamma = \frac{1}{3(3L_f + C)\sqrt{T}}.
\]
}

\begin{proofof}{Theorem~\ref{thmMainThm}}
Since $f$ has $L_f$-Lipschitz gradient, we know that 
\begin{align*}
	f (x_{t+1}) - f(x_t) &\leq -\gamma \cdot \inn{\nabla f(x_t)}{g_{\xi_t}(x_t) + \Delta_{\xi_t}(x_t)} + \frac{\gamma^2 L_f}{2} \| g_{\xi_t}(x_t) + \Delta_{\xi_t}(x_t)\|^2.
	\end{align*}
	Since the quantization is unbiased, implying $\E_Q [\Delta_{\xi}^{(Q)}(x)] = 0$, taking the expectation with respect to the quantization yields
	\begin{align*}
	&\E_Q [f(x_{t+1})] - \E_Q [f(x_t)] \\
	&\hspace{10mm}\leq - \gamma \E_Q \inn{\nabla f(x_t)}{g_{\xi_t}(x_t) + \tilde\Delta_{\xi_t}(x_t)} + \frac{3\gamma^2 L_f}{2} \left(  \| g_{\xi_t}(x_t) \|^2+\| \tilde\Delta_{\xi_t}(x_t)\|^2 + \E_Q\| \Delta_{\xi_t}^{(Q)}\|^2\right),
	\end{align*}
	where $\tilde\Delta_{\xi_t}(x_t) = (\Delta_{\xi_t}^{(a)}, \Delta_{\xi_t}^{(b)})$, and we used $(\alpha + \beta + \rho)^2 \leq 3\alpha^2 + 3\beta^2 + 3\rho^2$.
	
	Taking the expectation over $\xi_t$ (simplifying the notation of $\E_Q \E_{\xi}$ to simply $\E$), and recalling that $\E[g_{\xi_t}(x_t) ] = \nabla f(x_t)$, we can bound the first two terms of the RHS by
	\begin{align*}
	    &- \gamma \E \inn{\nabla f(x_t)}{g_{\xi_t}(x_t) + \tilde\Delta_{\xi_t}(x_t)} + \frac{3}{4}\gamma^2 L_f \E \| g_{\xi_t}(x_t) + \tilde\Delta_{\xi_t}(x_t)\|^2  \\
	    %&\leq -\gamma \E \| \nabla f(x_t)\|^2 - \gamma \E \inn{\nabla f(x_t)}{\tilde\Delta_{\xi_t}(x_t)} + \frac{3}{4}\gamma^2 L_f \E \| g_{\xi_t}(x_t) + \tilde\Delta_{\xi_t}(x_t)\|^2 \\ 
	&\leq -\frac{\gamma}{2} \E \|\nabla f(x_t)\|^2 + \frac{\gamma}{2} \E \|\tilde\Delta_{\xi_t}(x_t) \|^2 + \frac{3}{2}\gamma^2 L_f \E \left(\| g_{\xi_t}(x_t)\|^2 +  \|\tilde\Delta_{\xi_t}(x_t)\|^2\right)  \hspace{5mm} \{ \alpha \cdot \beta \leq \frac{1}{2}(\alpha^2+\beta^2)\}  \\
	 &\leq \left(-\frac{\gamma}{2}+3\gamma^2 L_f\right) \E \| \nabla f (x_t)\|^2 + \left( \frac{\gamma}{2} + \frac{3}{2}\gamma^2 L_f \right) \E \|\tilde\Delta_{\xi_t}(x_t)\|^2  + 3\gamma^2 L_f \sigma^2 \hspace{2mm} \{ \E \| g_{\xi_t} - \nabla f \|^2 \leq \sigma^2 \} \\
	 &\leq \left(-\frac{\gamma}{2}+3\gamma^2 L_f\right) \E \| \nabla f (x_t)\|^2  \\
	&\hspace{20mm} +\left( \frac{\gamma}{2} + \frac{3}{2}\gamma^2 L_f \right)(1+C_{a})^2 L_{f\circ b}^2 \E \|\delta_{\xi_t}(x_t)\|^2 + 3\gamma^2 L_f \sigma^2. \hspace{5mm} \{ \text{Lemma~\ref{lemAuxTwo}} \}
	\end{align*}
	Reorganizing the terms, summing over all $t\in [T]$ and dividing by $T$ yields
	\begin{align*}
	\gamma \cdot \left( \frac{1}{2} - 3\gamma L_f\right) \cdot \frac{1}{T} \sum_{t\in [T]} \E  \| \nabla f (x_t)\|^2 &\leq \frac{f(x_1) - \E[f(x_{T+1})]}{T} + \gamma \cdot C'' \cdot \frac{1}{T} \sum_{t\in [T]} \E \| \delta_{\xi_t}(x_t) \|^2 \\
	&\hspace{30mm}+ \gamma^2 L_f (3\sigma^2 +  \frac{3}{2} \cdot (c_QC_aC_{f\circ b})^2), 
	\end{align*}
	where 
	\begin{equation}\label{eqncThree}
	    C'' = \left(\frac12 + \frac{3}{2}\gamma L_f\right) (1+C_{a})^2 L_{f\circ b}^2 < (1+C_{a})^2 L_{f\circ b}^2,
	\end{equation}
	by the definition of $\gamma$. Applying Lemma~\ref{lemAuxTwo} and regrouping the terms now yields
	\begin{align*}
	    &\gamma \cdot \left(  \frac{1}{2} - 3\gamma L_f - \gamma^2 \cdot C' C'' \right)  \cdot \frac{1}{T} \sum_{t\in [T]} \| \nabla f (x_t)\|^2 \\
	    &\hspace{20mm}\leq \frac{f(x_1) - \E[f(x_{T+1})]}{T} + \gamma^2 \left( \gamma \cdot C'C''  + 3L_{f} \right) \cdot  (\sigma^2 + (c_Q C_a C_{f\circ b})^2).
	\end{align*}
	Noting that $C'C'' < C^2$ and recalling that $\gamma C < 1/3$ and $\gamma L_f < 1/9$, we get
	\begin{align*}
	    &\gamma \cdot \left(  \frac{1}{2} - \gamma (3L_f +C) \right)  \cdot \frac{1}{T} \sum_{t\in [T]} \| \nabla f_{\xi_t}(x_t)\|^2 \\
	    &\hspace{20mm}\leq \frac{f(x_1) - \E[f(x_{T+1})]}{T} + \gamma^2 \left(3L_{f} + C\right) \cdot  (\sigma^2 + (c_Q C_a C_{f\circ b})^2).
	\end{align*}
	Since $\gamma\cdot(3L_f + C) = \frac{1}{3\sqrt{T}} \leq  \frac{1}{3}$, the LHS coefficent is at least $\gamma/6$, so dividing by $\gamma /6$ now yields the claim by substituting $\gamma$ with $\frac{1}{3(3L_f + C)\sqrt{T}}$.
\end{proofof}

\subsection{Theoretical analysis when $K>2$}
In this section we sketch how one can generalize the already provided theoretical analysis for the $K=2$ case.

Instead of Machines $a$ and $b$, we now suppose that we have a stack $a_1$, \ldots, $a_K$ of $K$ Machines such that every pair $a_i, a_{i+1}$ communicates a compressed message, denoted by $m_i$. We simplify the notation of the complete model by $x=(x^{(1)}, \ldots, x^{(K)})$, and, for a sample $\xi$, further denote
\begin{align*}
\overline{a}^{(i)} (\xi, x_t) &= a_{i} (a_{i-1}(\ldots ( a_1(\xi, x_{t}^{(1)}), x_{t}^{(2)}),\ldots x_{t}^{(i)})),\\
\overline{m}^{(i)} (\xi, x_t)&= m_{i} (m_{i-1}(\ldots ( m_1(\xi, x_{t}^{(1)}), x_{t}^{(2)}),\ldots x_{t}^{(i)})),
\end{align*}
for $i\in[K]$. As in Algorithm~\ref{alg:SHwT},
for iteration $t$ and the data sample $\xi_t$, 
\textit{if} it is the first time that
$\xi_t$ is sampled, 
Machine $a_i$ communicates to Machine $a_{i+1}$ the full
precision activations without any compression:
$m_i(\xi_t) = \overline{a}^{(i)}(\xi_t,x_t)$.
If $\xi_t$ has been
sampled in previous iterations,
Machine $a_i$ communicates a compressed 
version:
\[
Q(a_i(\overline{m}^{(i-1)}(\xi_t), x_t^{(i)}) - m_i (\xi_t)),
\]
where $m_i(\xi_t)$ was the previous 
message, stored in the local buffer.
Both machines
then update this local buffer:
\vspace{-0.3em}
\[
m_i(\xi_t) \leftarrow m_i(\xi_t) + Q(a_i(\overline{m}^{(i-1)}(\xi_t), x_t^{(i)}) - m_i (\xi_t)).
\]
Machine $a_{i+1}$ then uses 
$m_i(\xi_t)$ as the forward activations. Upon receiving backward gradients from Machine $a_{i+2}$, it 
computes backward gradients,
and communicates a quantized 
version of the backward gradient to 
Machine $a_i$. We use
\[
\delta_{\xi}^{(i)} = \overline{a}^{(i)} (\xi, x_t) - \overline{m}^{(i)} (\xi, x_t)
\]
to denote the \textit{message error} of $i$-th machine in
sending the activations (accumulated also through messages in previous pairs), and denote the \textit{total message error} by $\delta_\xi = (\delta_{\xi}^{(1)}, \ldots, \delta_{\xi}^{(K-1)})$.

{\bf Update Rules.} 
We can now generalize the update rule for $a$ and $b$ to 
\begin{align*}
    x_{t+1}^{(K)} &= x_{t}^{(K)} - \gamma \cdot \nabla_{x^{(K)}} (f\circ a_K)\vert_{(\overline{m}^{(K-1)}(\xi_t),x_{t}^{(K)})}, \\
    x_{t+1}^{(i)} &= x_{t}^{(i)} - \gamma \cdot Q(\nabla_{a_i} (f\circ a_K \circ \ldots \circ a_{i+1})\vert_{(\overline{m}^{(i)}(\xi_t),x_{t}^{(i+1)}, \ldots, x_{t}^{(K)})}) \cdot \nabla_{x^{(i)}} a_i\vert_{(\overline{m}^{(i-1)}(\xi_t), x_{t}^{(i)})},
\end{align*}
for $i=1,\ldots, K-1$, where $\gamma$ is the learning rate. We can rephrase the update rule as
\[
x_{t+1} = x_{t} - \gamma \cdot \left( g_{\xi} (x_t) + \Delta_{\xi} (x_t) \right),
\]
where 
$g_{\xi} (x_t)$ is the stochastic gradient and 
$\Delta_{\xi} (x_t)$ is the \textit{total gradient error} 
introduced by communication compression.
We have 
$\Delta_{\xi} = (\Delta_{\xi}^{(1)} + \Delta_{\xi}^{(Q,1)},\ \ldots \ , \Delta_{\xi}^{(K-1)} + \Delta_{\xi}^{(Q,K-1)}, \Delta_{\xi}^{(K)})$, given by:
\begin{align*}
    \Delta_{\xi}^{(Q,i)} (x_t) &= Q(\nabla_{a_i} (f\circ a_K \circ \ldots \circ a_{i+1})\vert_{(\overline{m}^{(i)}(\xi_t),x_{t}^{(i+1)}, \ldots, x_{t}^{(K)})}) \cdot \nabla_{x^{(i)}} a_i\vert_{(\overline{m}^{(i-1)}(\xi_t), x_{t}^{(i)})} \\
    &\hspace{10mm}-\nabla_{a_i} (f\circ a_K \circ \ldots \circ a_{i+1})\vert_{(\overline{m}^{(i)}(\xi_t),x_{t}^{(i+1)}, \ldots, x_{t}^{(K)})} \cdot \nabla_{x^{(i)}} a_i\vert_{(\overline{m}^{(i-1)}(\xi_t), x_{t}^{(i)})},\\
    \Delta_{\xi}^{(i)} (x_t) &= \nabla_{a_i} (f\circ a_K \circ \ldots \circ a_{i+1})\vert_{(\overline{m}^{(i)}(\xi_t),x_{t}^{(i+1)}, \ldots, x_{t}^{(K)})} \cdot \nabla_{x^{(i)}} a_i\vert_{(\overline{m}^{(i-1)}(\xi_t), x_{t}^{(i)})} \\
    &\hspace{10mm}- \nabla_{a_i} (f\circ a_K \circ \ldots \circ a_{i+1})\vert_{(\overline{a}^{(i)}(\xi_t),x_{t}^{(i+1)}, \ldots, x_{t}^{(K)})} \cdot \nabla_{x^{(i)}} a_i\vert_{(\overline{a}^{(i-1)}(\xi_t), x_{t}^{(i)})},\\
    \Delta_{\xi}^{(K)} (x_t) &= \nabla_{x^{(K)}} (f\circ a_K)\vert_{(\overline{m}^{(K-1)}(\xi_t),x_{t}^{(K)})} - \nabla_{x^{(K)}} (f\circ a_K)\vert_{(\overline{a}^{(K-1)}(\xi_t),x_{t}^{(K)})},
\end{align*}
for all $i=1,\ldots,K-1$.

\paragraph{Generalized Assumptions.} In order to state the analogue of Theorem~\ref{thmMainThm} for $K>2$, we need to define the corresponding assumptions with respect to Lipschitz properties (whereas Assumption GA2 is here only for completeness, being the same as A2). 

\begin{itemize}[leftmargin=*]
    \item \textbf{(GA1: Lipschitz assumptions)} We assume that 
    \begin{itemize}
        \item $f$ had $L_f$-Lipschitz gradient,
        \item $f\circ a_K \circ \ldots \circ a_{i+1}$ has $L_{f\circ a_K \circ \ldots \circ a_{i+1}}$-Lipschitz gradient, and has gradient bounded by $C_{f\circ a_K \circ \ldots \circ a_{i+1}}$ for all $i=1,\ldots, K-1$,
        \item $a_i$ is $\ell_{a_i}$-Lipschitz, and has gradient bounded by $C_{a_i}$, for all $i=1,\ldots, K$.
    \end{itemize}
    \item \textbf{(GA2: SGD assumptions)} We assume that the stochastic gradient $g_\xi$ is unbiased, i.e. $\E_\xi [g_\xi (x)] = \nabla f(x)$, for all $x$, and with bounded variance, i.e. $\E_{\xi} \| g_\xi (x) - \nabla f(x) \|^2 \leq \sigma^2$, for all $x$.
\end{itemize}

We have the following analogue of Theorem~\ref{thmMainThm}.
\begin{thm}\label{thmMainGenThm}
Suppose that Assumptions GA1, GA2 hold, and let 
\[
\tilde C = \sqrt{\sum_{i=1}^{K-1} C_{a_i}^2 C_{f\circ a_K \circ \ldots \circ a_{i+1}}^2}.
\]
Consider 
an unbiased quantization function
$Q(x)$ which satisfies that there exists $ c_Q < \sqrt{1/2}$ such that $\E \| x-Q(x) \| \leq c_Q \|x\|$, for all $x$. Then there exists a constant $C$ that depends only on the constants defined above and on $\sqrt{K}$ and $N$, such that for
the learning rate $\gamma$ proportional to  $\frac{1}{(C+L_f)\sqrt{T}}$, after performing $T$ updates one has
\begin{equation}\label{eqnMainGenResult}
    \frac{1}{T} \sum_{t\in [T]} \E \| \nabla f(x_t)\|^2 \lesssim  \frac{(C+L_f)(f(x_1) - f^*)}{\sqrt{T}} + \frac{\sigma^2 + (c_{Q}\tilde C)^2}{\sqrt{T}}.
\end{equation}
\end{thm}

Instead of performing a tedious job of rewriting the proof of the $K=2$ case with inherently more constant-chasing, we will simply sketch the differences with respect to the proof of Theorem~\ref{thmMainThm}. First we note that, having analogous assumptions as in the case when $K=2$, we can easily prove the following analogue of Lemma~\ref{lemAuxOne}.
\begin{lem}\label{lemAuxGenOne}
Let 
\[
\Delta_{\xi}^{(Q)} = (\Delta_{\xi}^{(Q,1)}, \ldots, \Delta_{\xi}^{(Q,K-1)}).
\] 
For every sample $\xi$, one has 
\[
\|\Delta_{\xi}^{(Q,i)} (x)\| \leq c_Q C_{a_i} C_{f\circ a_K \circ \ldots \circ a_{i+1}},\hspace{10mm} i=1,\ldots, K-1,
\]
\[
\|\Delta_{\xi}^{(i)} (x)\| \leq C_{a_i} L_{f\circ a_K \circ \ldots \circ a_{i+1}} \| \delta_{\xi}^{(i)}\| + C_{f\circ a_K \circ \ldots \circ a_{i+1}} L_{a_i}\|\delta_{\xi}^{(i-1)}\|,\hspace{10mm} i=1,\ldots, K-1,
\]
and 
\[
\|\Delta_{\xi}^{(K)} (x)\| \leq L_{f\circ a_K}  \| \delta_{\xi}^{(K-1)}\|,
\]
implying $\|\Delta_{\xi}^{(Q)}\| \leq \tilde C c_Q$.
\end{lem}
Comparing this with Lemma~\ref{lemAuxOne}, we see that for $\tilde \Delta_\xi = (\Delta_{\xi}^{(1)},\ldots, \Delta_{\xi}^{(K)})$, we now have an additional term that depends on $\delta_{\xi}^{(i-1)}$. However, since in the proof of Theorem~\ref{thmMainThm} we only rely on $\|\Delta_{\xi}\|^2$, we can proceed even with a crude bound
\begin{align*}
&\| \tilde \Delta_{\xi} \|^2 = \sum_{i=1}^{K} \| \Delta_{\xi}^{(i)}\|^2 \\
&\leq (1+2C_{a_{K-1}}^2)L_{f\circ a_K}^2 \| \delta_{\xi}^{(K-1)}\|^2 + 2 \sum_{i=1}^{K-2} (C_{a_{K-2}}^2L_{f\circ a_K \circ \ldots \circ a_{i+1}}^2 + C_{f\circ a_K \circ \ldots \circ a_{i+2}}^2 L_{a_{i+1}}^2) \|\delta_{\xi}^{(i)}\|^2 \\
&\leq 2K \cdot \underbrace{\max \left\{ (1+2C_{a_{K-1}}^2)L_{f\circ a_K}^2, \max_{i\in[K-2]} \{ C_{a_{K-2}}^2L_{f\circ a_K \circ \ldots \circ a_{i+1}}^2 + C_{f\circ a_K \circ \ldots \circ a_{i+2}}^2 L_{a_{i+1}}^2\} \right\}}_{C_1} \cdot \| \delta_\xi\|^2.
\end{align*}

Mimicking closely the steps of Lemma~\ref{lemAuxTwo}, separately for each $i$ and the summing over all $i$ via $\| \ell_a \|^2 = \sum_{i=1}^{K} \|\ell_{a_i}\|^2$, one can straightforwardly yield the following analogue of Lemma~\ref{lemAuxTwo}.
\begin{lem}\label{lemAuxGenTwo}
For $C' = K \cdot \frac{36c_{Q}^2 \|l_{a}\|^2 N^2 C_1}{1-2c_{Q}^2}$, where $l_{a} = (l_{a_1},\ldots, l_{a_K})$, one has
\[
\frac{1}{T} \sum_{t\in [T]} \E \| \delta_{\xi_t}(x_t) \|^2 \leq C' \gamma^2 \cdot \left( \frac{1}{T} \sum_{t\in [T]} \E \| \nabla f (x_t)\|^2 + \sigma^2 + (c_Q \tilde C)^2\right).
\]
\end{lem}

Theorem~\ref{thmMainGenThm} can now be proved by repeating the steps of the proof of Theorem~\ref{thmMainThm}, carefully substituting Lemma~\ref{lemAuxOne} by Lemma~\ref{lemAuxGenOne}, and Lemma~\ref{lemAuxTwo} by Lemma~\ref{lemAuxGenTwo}.

\newpage

\section{Benchmark Dataset} \label{apd:data}

The data statistics can be found in Table \ref{tab:data}.
We fine-tune DeBERTa on QNLI and CoLA datasets.
The QNLI task is to determine whether the context sentence contains the answer to the question.
The CoLA task aims to detect whether a given sequence of tokens is a grammatical English sentence.
In addition, we fine-tune GPT2 on the WikiText2 training set. 
We also collect 30K arXiv abstracts in 2021 to fine-tune GPT2.
Neither corpus is included in the GPT2 pre-training data set.

\begin{table}[]
    \centering
    \begin{tabular}{llll}
    \toprule
        Dataset & \# labels & \# train samples & Task description \\
    \midrule
        QNLI    &  2        &    105K question-paragraph pairs & natural language inference \\
        CoLA    &  2        &    8.6K sentences & linguistic acceptability \\
        WikiText2 & -       &    2M tokens     & language modeling    \\
        arXiv     & -       &    7M tokens     & language modeling    \\
    \bottomrule \\
    \end{tabular}
    \caption{Dataset statistics}
    \label{tab:data}
\end{table}

\section{Training Task Setup and Hyper-parameter Tuning } \label{apd:setup}

\textbf{Sequence Classification.}
We use the AdamW optimizer to fine-tune the model for 10 epochs.
Specifically, for QNLI, we set the learning rate to 3.0e-6, 
learning rate warm-up steps to 1000, max sequence length to 256, the macro-batch size to 64 and micro-batch size to 8;
for CoLA, we set the learning rate to 2.5e-6, 
learning rate warm-up steps to 250, max sequence length to 128, the macro-batch size to 32 and micro-batch size to 8.
After the learning rate warm-up stage, we decay the learning rate linearly over the training epochs.

\textbf{Language Modeling.}
For both WikiText and arXiv datasets, we use AdamW optimizer with a learning rate of 5.0e-6.
We train the model for 10 epochs with a macro-batch size of 32 and a micro-batch size of 1.
The max sequence length is set to 1024 for both datasets.
After the learning rate warm-up stage, we decay the learning rate linearly over the training epochs.

% \textbf{Split Learning}
% We set 16 clients and use a Dirichlet distribution with concentration parameter 0.5 to synthesize non-identical datasets.
% Following the previously established work of split learning,
% in each communication round, we conduct local training for each client sequentially.
% and each client will train 3 epochs with its local data.
% We utilize SGD optimizer with momentum of 0.9, a batch size of 64, and a learning rate of 0.01.
% We decay the learning rate to its 10\% for every 20 communication rounds.

% The datasets are augmented with random cropping and flipping.
% To adapt to random cropping, we do the same cropping operation on the retrieved cached activation, and only update its non-cropped part.
% To adapt to random flipping, we maintain another activation cache copy for flipped images, and retrieve and update only the corresponding copy during training.

\section{Distributed View of \algname lgorithm}

Algorithm \ref{alg:SHwT2} shows a multi-node view of \algname.
For brevity, we omit the first warm-up epoch, 
where we conduct uncompressed training, and thus we update the previous messages by $m(\xi) \leftarrow a(\xi, x)$.

\begin{algorithm}[ht]
  \caption{\algname Algorithm}
  \label{alg:SHwT2}
\begin{algorithmic}
  \STATE {\bfseries Initialize:} $x_0$, learning rate $\gamma$, network $a$'s weights $x^{(a)}$, network $b$'s weights $x^{(b)}$, quantization function $Q$, the arrays of previous messages $m$, where networks $a$ and $b$ each maintain a copy of it.  \\
  
    \FOR{t = 1, \ldots, T}

    \STATE {(\bfseries on network $a$)}
        \STATE Randomly sample $\xi_t$ 
        \STATE $\Delta m(\xi_t) \leftarrow Q\big( a(\xi_t, x_{t}^{(a)}) - m(\xi_t)\big)$
        \STATE Update $m(\xi_t) \leftarrow m(\xi_t) + \Delta m(\xi_t)$
        \STATE Send $\Delta m(\xi_t)$ to network $b$
    
    \STATE {(\bfseries on network $b$)}
        \STATE Update $m(\xi_t) \leftarrow m(\xi_t) + \Delta m(\xi_t)$
        % \STATE Compute $\nabla_{x^{(b)}} (f\circ b)\vert_{m}$
        \STATE Update $x_{t+1}^{(b)} \leftarrow x_{t}^{(b)} - \gamma \cdot \nabla_{x^{(b)}} (f\circ b)\vert_{m}$
        \STATE Send $Q(\nabla_{a} (f\circ b)\vert_{m})$ to network $a$
    
    \STATE {(\bfseries on network $a$)}
        \STATE Update $x_{t+1}^{(a)} \leftarrow x_{t}^{(a)} - \gamma \cdot Q(\nabla_{a} (f\circ b)\vert_{m}) \cdot \nabla_{x^{(a)}} a$ 
    
    \ENDFOR
    \STATE {\bfseries Output:} $x = (x_{T}^{(a)}, x_{T}^{(b)})$\\    
\end{algorithmic}
\end{algorithm}

{
% \color{blue}

\section{Decentralized Training over Slow Network}

Decentralized training for large foundation models recently attracted intensive interests. 
Example projects include Learning@home \cite{ryabinin2020towards}, DeDLOC \cite{diskin2021distributed}, and Training Transformers Together \cite{borzunov2022training}. 
The goal of these projects is to enable a decentralized open-volunteering paradigm for foundation model training. 
As many geo-distributed users contribute their GPUs, these GPUs are often connected via slow networks. 
For example, \cite{diskin2021distributed} investigates a heterogeneous with bandwidths of 200Mbps, 100Mbps, and 50Mbps; 
\cite{borzunov2022training} advocates to train Transformer over the Internet with 10-100Mbps bandwidth; 
\cite{ryabinin2021swarm} considers network bandwidth less than 400Mbps. 

In this setting, communication compression is key to performance. However, when compressing activations, existing methods rely on direct quantization. This inspired our paper, which provides the first activation quantization method with rigorous theoretical guarantee and outperforms direct quantization.

\section{Discussion on Tensor Parallelism} 

We here discuss tensor parallelism \cite{shoeybi2019megatron}, and the potential adaptation of our algorithm to it.
In tensor parallelism, the activations are computed across different machines, and need to be aggregated. 
Therefore we need to compress activations both before and after allreduce to support tensor parallelism. 
Specifically, suppose $N$ machines conduct tensor parallelism, 
then the output activation is: 
\begin{equation}
A = A_1+A_2+ ... +A_N,
\end{equation}
and we need to compress communication twice:
\begin{equation}
A_Q = Q[ Q(A_1)+Q(A_2)+…+Q(A_N) ].
\end{equation}
We believe that delta compensation could be applied to all $Q(-)$, similar to how previous work handles gradient compression (e.g. Eq. 3 and 4 in \cite{tang2019doublesqueeze}). 
However, this requires careful further studies, both empirically and theoretically. 
We leave activation compression for tensor parallelism as future work.

\section{Limitation and Potential Future Direction}

\paragraph{Additional Storage.}
Our algorithm trades storage for communication. 
Fortunately, we find this is a reasonable trade-off in our settings. In the following we show that we can offload the activations to SSD and hide it within the GPU computation of other data examples.

We compare the throughput under the bandwidth of 10Gb/s.
\texttt{FP32} achieves a throughput of 3.8 seqs/s, 
while \algname, either offloading activations to host memory or SSDs, achieves 4.0 seqs/s.
Considering the similar training throughput of the above three settings,
we show that the overhead of offloading to SSD can be successfully hided in GPU computation.
In particular, our largest experiment in this paper requires 172GB storage per machine, which 
even with a 10x larger dataset can be easily offloaded to SSDs.
For much larger datasets (e.g. 100x), we can use data parallelism to reduce the storage requirement for each machine. 
For an even larger dataset, our algorithm might not support it well. This then requires further studies.

\paragraph{Online Learning.}
Our proposal relies on iterating over multiple epochs, which is a common setting. 
We understand our current algorithm has limitations in the settings such as online learning. 
In the following, we provide a potential solution (to both storage requirement and limitation in online learning) – relaxing \algname by clustering activations and storing only the centers of the clusters. 

Recall that \algname compresses the ``delta” (difference between activations from two epochs) and thereby needs to store activations from the previous epoch. In the relaxed \algname, we could use algorithms like clustering or locality sensitive hashing to partition the activations and then we only store the ``centers” of each partition/cluster. When computing the ``delta”, we can first identify which partition/cluster the current activation belongs to and retrieve the corresponding ``center”. Then ``delta” = activation - ``center”. This will potentially help address storage and online learning limitations. We will explore this in future work.

}

\section{Additional Results}

We provide additional experimental results.
Specifically, we show: 
\setlist{nolistsep}
\begin{itemize}
    \setlength\itemsep{0em}
    \item the convergence results with standard deviation; 
    \item the training throughput for different dataset settings;
    % \item {\color{blue} additional analysis on generation models};
    \item {
    % \color{blue} 
    the numerical stability of training from scratch};
    \item {
    % \color{blue} 
    the training results under FP16 precision};
    \item the robustness of \algname under different hyperparameter settings;
    \item the effectiveness of \algname in the split learning scenario.
\end{itemize}

\subsection{Convergence Results with Standard Deviation}

\begin{figure}
    \centering
    \begin{subfigure}[b]{0.245\linewidth}
        \centering
        \includegraphics[width=\linewidth]{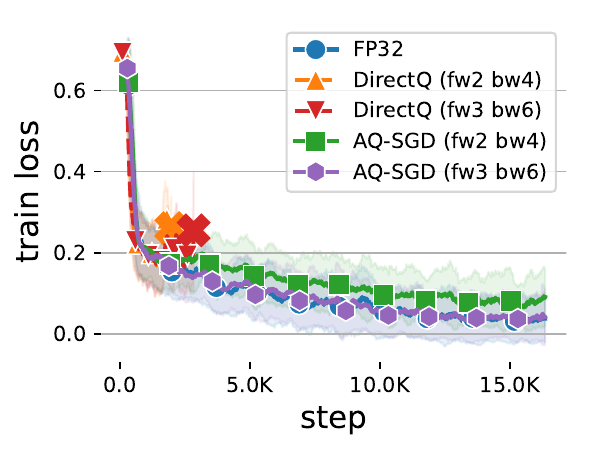}
        \caption{QNLI, DeBERTa-1.5B}
        % \label{fig:exp_qnli}
    \end{subfigure}
    \hfill
    \begin{subfigure}[b]{0.245\linewidth}
        \centering
        \includegraphics[width=\linewidth]{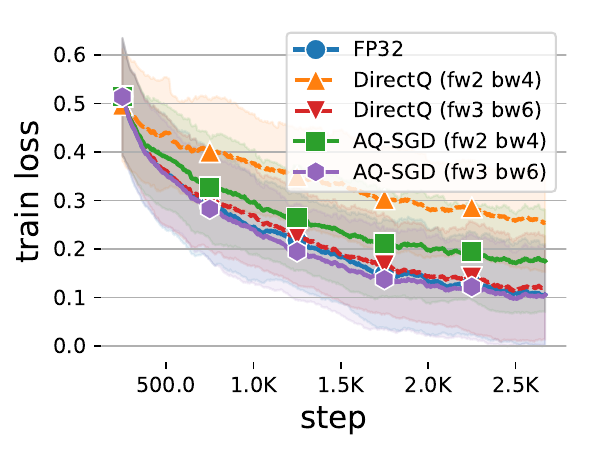}
        \caption{CoLA, DeBERTa-1.5B}
        % \label{fig:exp_cola}
    \end{subfigure}
    \hfill
    \begin{subfigure}[b]{0.245\linewidth}
        \centering
        \includegraphics[width=\linewidth]{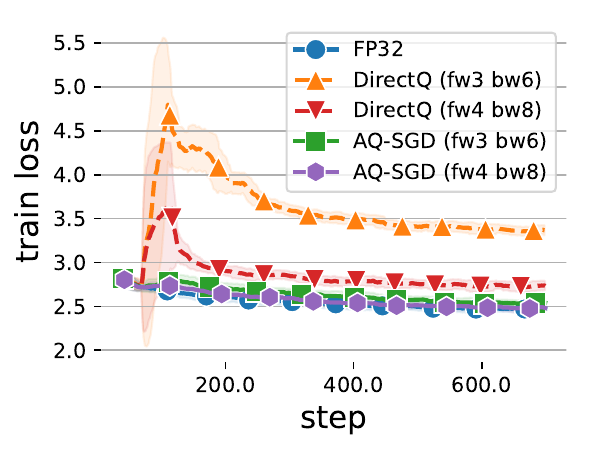}
        \caption{WikiText2, GPT2-1.5B}
        % \label{fig:exp_wiki}
    \end{subfigure}
    \hfill
    \begin{subfigure}[b]{0.245\linewidth}
        \centering
        \includegraphics[width=\linewidth]{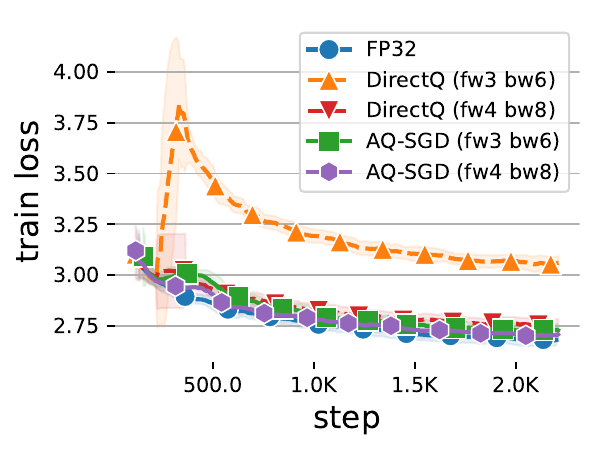}
        \caption{arXiv, GPT2-1.5B}
        % \label{fig:exp_arxiv}
    \end{subfigure}
       \caption{Convergence (loss vs. \# stpes) of different approaches. $\boldsymbol{\times}$ represents divergence.}
       \label{fig:exp_convs_error_bars}
\end{figure}

In the main content, we show the convergence performance of different approaches.
We repeated each experiment three times to ensure reproducibility.
We calculate the moving averages of these convergence curves 
and then average the results of repeated experiments.
% (the same average strategy was used in Figure~\ref{fig:exp_convs}).
We visualize (shaded areas) the moving standard deviation in all repeated experiments 
in Figure~\ref{fig:exp_convs_error_bars}.
Overall, we observe consistent results for all datasets.

\subsection{Throughput under Different Dataset Settings}

We show the training throughput under different dataset settings in Figure~\ref{tab:efficiency2}.
In general, the observation is similar to that of the main content:
our approach maintains similar throughput even when the network is 100$\times$ slower (from 10Gbps to 100Mbps).
WikiText2 and arXiv have essentially the same throughput results, since we use the same training settings for them.

\begin{table}[t!]
    \centering
    \scriptsize
    \caption{Training Throughput.}
    \setlength\tabcolsep{4 pt}
    \begin{tabular}{@{}rcccccc@{}}
    \toprule
        ~
         
        & \multicolumn{3}{c}{DeBERTa-1.5B, QNLI}  &    
        \multicolumn{3}{c}{GPT2-1.5B, WikiText2} \\  \cmidrule(lr){2-4} \cmidrule(lr){5-7}
        \begin{tabular}{@{}c@{}}
            Network \\ Bandwidth
        \end{tabular}
        & \texttt{FP32} & 
        \begin{tabular}{@{}c@{}}
            \texttt{DirectQ} \\
            fw2 bw4 / fw3 bw6 
        \end{tabular}
        & \begin{tabular}{@{}c@{}}
            \algname \\
            fw2 bw4 / fw3 bw6 
        \end{tabular}   
        & \texttt{FP32} & 
        \begin{tabular}{@{}c@{}}
            \texttt{DirectQ} \\
            fw3 bw6 / fw4 bw8 
        \end{tabular}
        & \begin{tabular}{@{}c@{}}
            \algname \\
            fw3 bw6 / fw4 bw8 
        \end{tabular}  \\
    \midrule             
        10 Gbps  &12.9$_{\pm0.02}$   & 13.6$_{\pm0.02}$ / 13.6$_{\pm0.02}$ & 13.6$_{\pm0.02}$ / 13.5$_{\pm0.02}$   
                & 3.8$_{\pm0.01}$   &  4.0$_{\pm0.01}$ /  4.1$_{\pm0.01}$ &  4.0$_{\pm0.01}$ /  4.0$_{\pm0.01}$   \\
        1 Gbps   & 9.6$_{\pm0.02}$   & 13.3$_{\pm0.02}$ / 13.1$_{\pm0.02}$ & 13.3$_{\pm0.02}$ / 13.0$_{\pm0.02}$   
                & 3.2$_{\pm0.01}$   &  4.0$_{\pm0.01}$ /  4.0$_{\pm0.01}$ &  4.0$_{\pm0.01}$ /  3.9$_{\pm0.01}$   \\
        500 Mbps & 6.2$_{\pm0.03}$   & 13.0$_{\pm0.03}$ / 12.6$_{\pm0.03}$ & 12.9$_{\pm0.03}$ / 12.5$_{\pm0.03}$   
                & 2.7$_{\pm0.02}$   &  3.9$_{\pm0.01}$ /  3.9$_{\pm0.01}$ &  3.9$_{\pm0.01}$ /  3.9$_{\pm0.01}$   \\
        300 Mbps & 4.4$_{\pm0.04}$   & 12.5$_{\pm0.02}$ / 11.9$_{\pm0.03}$ & 12.4$_{\pm0.03}$ / 11.8$_{\pm0.03}$   
                & 1.8$_{\pm0.02}$   &  3.9$_{\pm0.01}$ /  3.8$_{\pm0.01}$ &  3.8$_{\pm0.01}$ /  3.8$_{\pm0.01}$  \\
        100 Mbps & 1.6$_{\pm0.04}$   & 10.7$_{\pm0.03}$ / \ \ 9.4$_{\pm0.03}$ & 10.6$_{\pm0.03}$ / \ \ 9.1$_{\pm0.03}$   
                & 0.5$_{\pm0.02}$   &  3.5$_{\pm0.02}$ /  3.0$_{\pm0.02}$ &  3.4$_{\pm0.01}$ /  3.0$_{\pm0.02}$  \\
    \bottomrule \\
    \end{tabular}
    
    \begin{tabular}{@{}rcccccc@{}}
    \toprule
        ~
         
        & \multicolumn{3}{c}{DeBERTa-1.5B, CoLA}  &
        \multicolumn{3}{c}{GPT2-1.5B, arXiv} \\  \cmidrule(lr){2-4} \cmidrule(lr){5-7}
        \begin{tabular}{@{}c@{}}
            Network \\ Bandwidth
        \end{tabular}
        & \texttt{FP32} & 
        \begin{tabular}{@{}c@{}}
            \texttt{DirectQ} \\
            fw2 bw4 / fw3 bw6 
        \end{tabular}
        & \begin{tabular}{@{}c@{}}
            \algname \\
            fw2 bw4 / fw3 bw6 
        \end{tabular}   
        & \texttt{FP32} & 
        \begin{tabular}{@{}c@{}}
            \texttt{DirectQ} \\
            fw3 bw6 / fw4 bw8 
        \end{tabular}
        & \begin{tabular}{@{}c@{}}
            \algname \\
            fw3 bw6 / fw4 bw8 
        \end{tabular}  \\
    \midrule             
        10 Gbps  &17.1$_{\pm0.03}$   & 18.0$_{\pm0.03}$ / 17.9$_{\pm0.03}$ & 17.9$_{\pm0.03}$ / 17.8$_{\pm0.03}$   
                & 3.8$_{\pm0.01}$   &  4.0$_{\pm0.01}$ /  4.1$_{\pm0.01}$ &  4.0$_{\pm0.01}$ /  4.0$_{\pm0.01}$   \\
        1 Gbps   &12.2$_{\pm0.03}$   & 17.4$_{\pm0.02}$ / 17.1$_{\pm0.02}$ & 17.3$_{\pm0.02}$ / 16.9$_{\pm0.02}$   
                & 3.2$_{\pm0.01}$   &  4.0$_{\pm0.01}$ /  4.0$_{\pm0.01}$ &  4.0$_{\pm0.01}$ /  3.9$_{\pm0.01}$   \\
        500 Mbps &8.9$_{\pm0.03}$   & 16.7$_{\pm0.03}$ / 16.2$_{\pm0.03}$ & 16.7$_{\pm0.03}$ / 16.1$_{\pm0.03}$   
                & 2.7$_{\pm0.02}$   &  3.9$_{\pm0.01}$ /  3.9$_{\pm0.01}$ &  3.9$_{\pm0.01}$ /  3.9$_{\pm0.01}$   \\
        300 Mbps &6.0$_{\pm0.04}$   & 16.1$_{\pm0.03}$ / 15.2$_{\pm0.03}$ & 16.0$_{\pm0.03}$ / 15.1$_{\pm0.03}$   
                & 1.8$_{\pm0.02}$   &  3.9$_{\pm0.01}$ /  3.8$_{\pm0.01}$ &  3.8$_{\pm0.01}$ /  3.8$_{\pm0.01}$  \\
        100 Mbps &2.2$_{\pm0.04}$   & 13.1$_{\pm0.03}$ / 11.5$_{\pm0.03}$ & 13.1$_{\pm0.03}$ / 11.3$_{\pm0.03}$   
                & 0.5$_{\pm0.03}$   &  3.5$_{\pm0.01}$ /  3.0$_{\pm0.02}$ &  3.4$_{\pm0.01}$ /  3.0$_{\pm0.01}$  \\
    \bottomrule \\
    \end{tabular}
    
    \label{tab:efficiency2}
\end{table}

{
% \color{blue}

\subsection{Results of Training from Scratch} \label{sec:scratch}

\begin{figure}
    \centering
    % \captionsetup{labelfont={color=blue}}
    \begin{subfigure}[b]{0.433\linewidth}
        \centering
        \includegraphics[width=\linewidth]{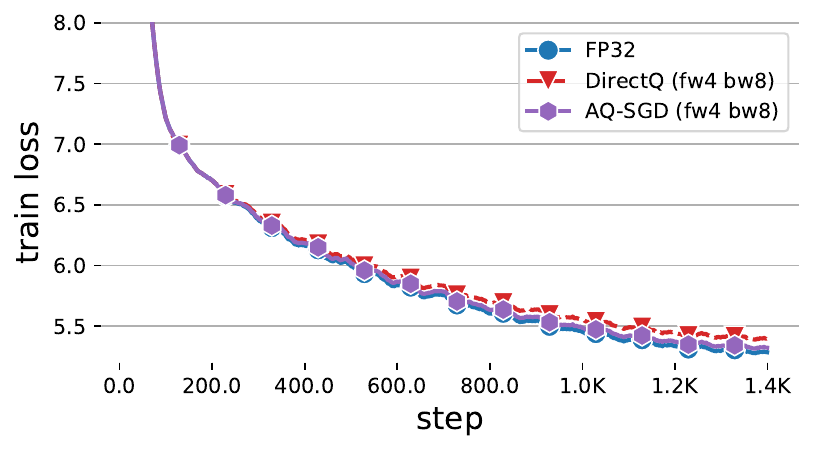}
        \caption{
        % \color{blue}
        Wiki from scratch}
        % \label{fig:}
    \end{subfigure}
    % \hfill
    \quad\quad
    \begin{subfigure}[b]{0.433\linewidth}
        \centering
        \includegraphics[width=\linewidth]{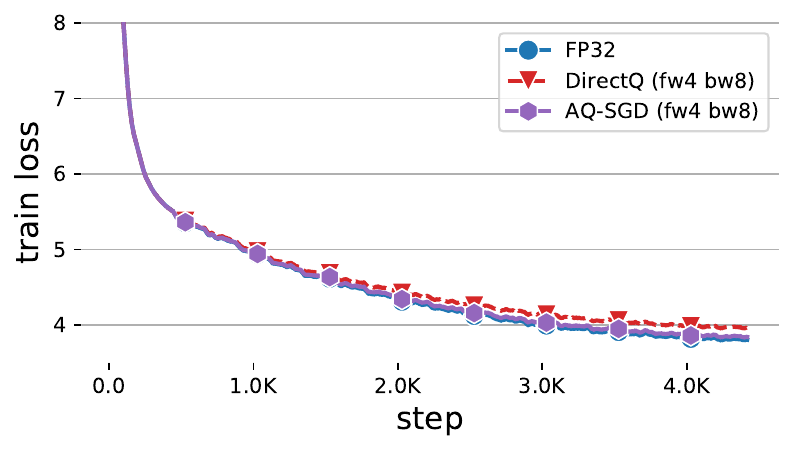}
        \caption{
        % \color{blue} 
        arXiv from scratch}
        % \label{fig:}
    \end{subfigure}
      \caption{
      % \color{blue} 
      Convergence results of training from scratch.}
      \label{fig:scratch}
\end{figure}

We investigate the numerical stability of \algname by showing the convergence result of training from scratch,
where the model parameters are randomly initialized.
We train WikiText and arXiv datasets for 20 epochs and use the first 10\% of steps as warm-up, respectively. 
As shown in Figure \ref{sec:scratch}, we can see that \algname converges almost as fast as \texttt{FP32} when training from scratch, 
which indicates our approach is robust enough even when the model is far from the converged state. 
In contrast, the curve of \texttt{DirectQ} becomes flatter in the late training stage, showing a clear gap with \texttt{FP32}.
}

{
% \color{blue}
\subsection{Results of Training under FP16} \label{sec:fp16}

\begin{figure}
    \centering
    % \captionsetup{labelfont={color=blue}}
    \begin{subfigure}[b]{0.245\linewidth}
        \centering
        \includegraphics[width=\linewidth]{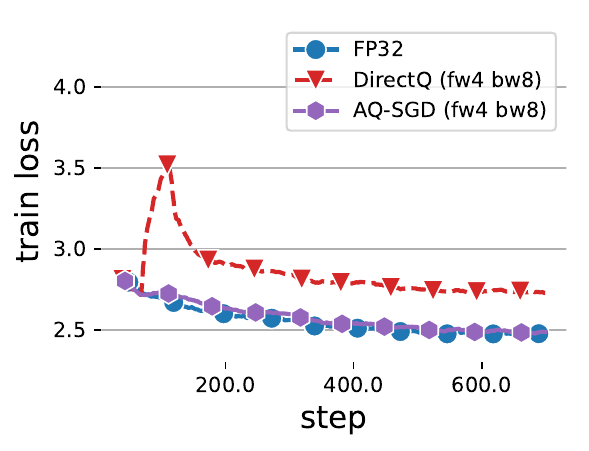}
        \caption{
        % \color{blue} 
        Wiki, FP32}
        % \label{fig:}
    \end{subfigure}
    \hfill
    \begin{subfigure}[b]{0.245\linewidth}
        \centering
        \includegraphics[width=\linewidth]{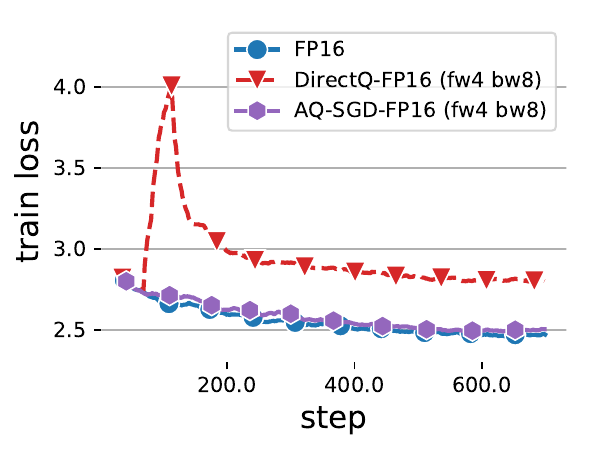}
        \caption{
        % \color{blue} 
        Wiki, FP16}
        % \label{fig:}
    \end{subfigure}
    \hfill
    \begin{subfigure}[b]{0.245\linewidth}
        \centering
        \includegraphics[width=\linewidth]{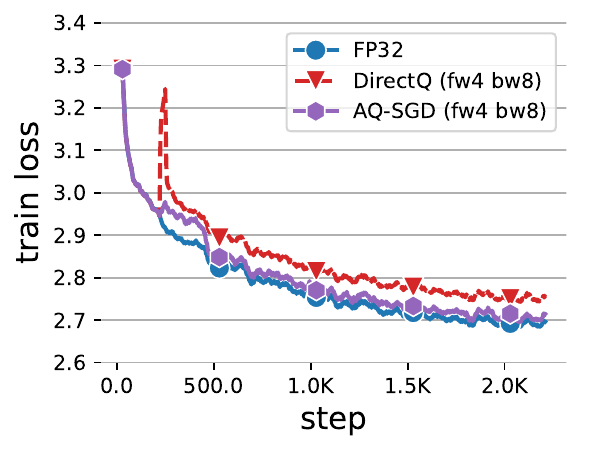}
        \caption{
        % \color{blue} 
        arXiv, FP32}
        % \label{fig:}
    \end{subfigure}
    \hfill
    \begin{subfigure}[b]{0.245\linewidth}
        \centering
        \includegraphics[width=\linewidth]{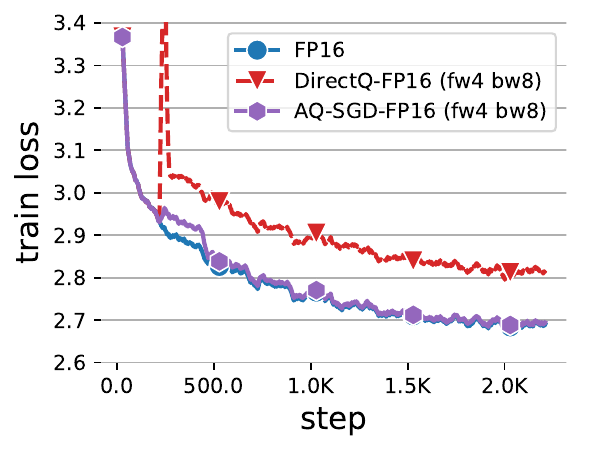}
        \caption{
        % \color{blue} 
        arXiv, FP16}
        % \label{fig:}
    \end{subfigure}
    
    \caption{
    % \color{blue} 
    Comparison of fine-tuning under FP32 and FP16.}
    \label{fig:fp32_fp16}
\end{figure}

To investigate the convergence performance of \algname under low-precision training, 
we here show the results of FP16 training, and compare it with FP32 training.
Figure \ref{fig:fp32_fp16} compares the results under FP32 and FP16.
In general, the convergence curves are consistent with the FP32 case.
This confirms the effectiveness of \algname when the activation is already in low precision.
}

\subsection{Hyper-parameter Sensitivity} \label{sec:n_cut}

Here we demonstrate the robustness of our method in various settings.
% We show provide results with different configurations here.
For fast validation, we focus on evaluating DeBERTa-v3-base\footnote{\url{https://huggingface.co/microsoft/deberta-v3-base}}
on QNIL and CoLA datasets.
We by default use $K=4$ devices for pipeline parallel training,
2 bits for forward activation, and 4 bits for backward gradients (\texttt{fw2 bw4}).

\begin{figure}
    \centering
    
    \begin{subfigure}[b]{0.245\linewidth}
        \centering
        \includegraphics[width=\linewidth]{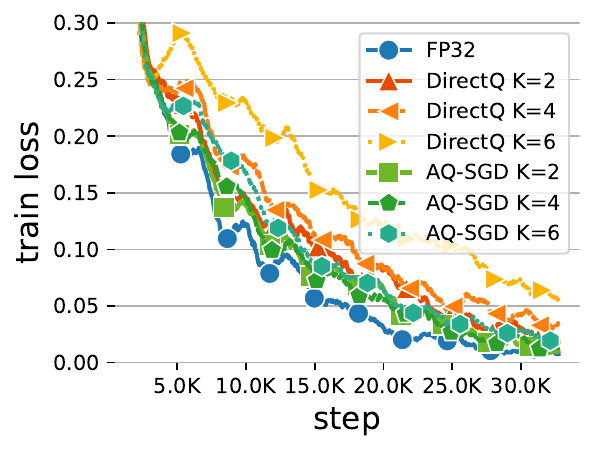}
        \caption{QNLI, $K$ stages}
        \label{fig:exp_n_cut_qnli}
    \end{subfigure}
    \hfill
    \begin{subfigure}[b]{0.245\linewidth}
        \centering
        \includegraphics[width=\linewidth]{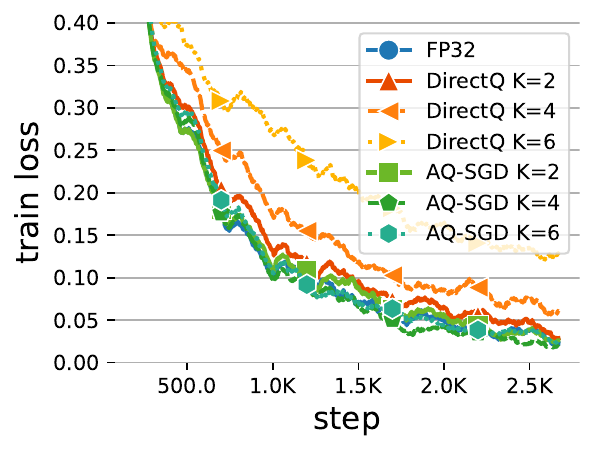}
        \caption{CoLA, $K$ stages}
        \label{fig:exp_n_cut_cola}
    \end{subfigure}
    \hfill
    \begin{subfigure}[b]{0.245\linewidth}
        \centering
        \includegraphics[width=\linewidth]{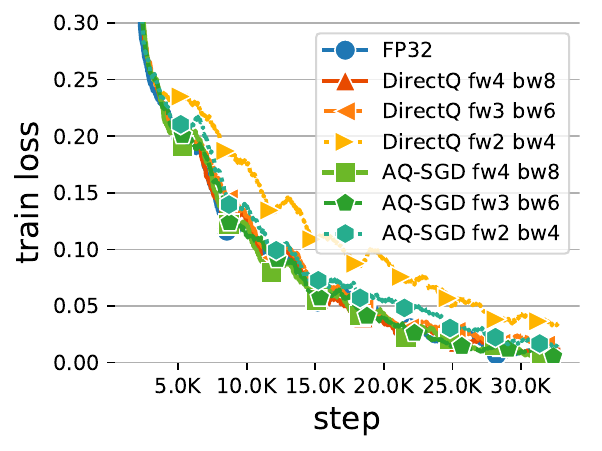}
        \caption{QNLI, $n$ bits}
        \label{fig:exp_n_bits_qnli}
    \end{subfigure}
    \hfill
    \begin{subfigure}[b]{0.245\linewidth}
        \centering
        \includegraphics[width=\linewidth]{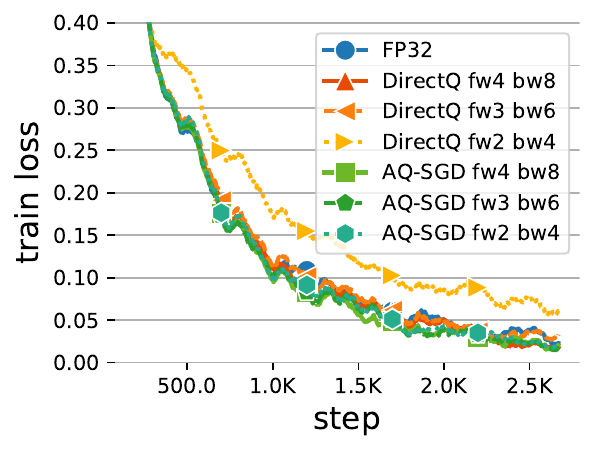}
        \caption{CoLA, $n$ bits}
        \label{fig:exp_n_bits_cola}
    \end{subfigure}

    \begin{subfigure}[b]{0.245\linewidth}
        \centering
        \includegraphics[width=\linewidth]{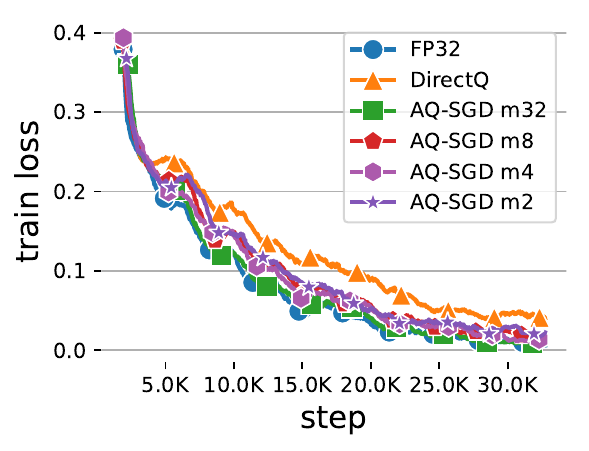}
        \caption{QNLI, low-bit $m$}
        \label{fig:exp_n_bits_cache_qnli}
    \end{subfigure}
    \hfill
    \begin{subfigure}[b]{0.245\linewidth}
        \centering
        \includegraphics[width=\linewidth]{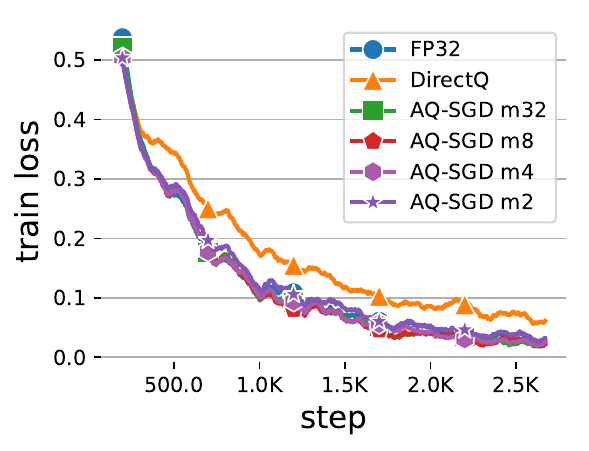}
        \caption{CoLA, low-bit $m$}
        \label{fig:exp_n_bits_cache_cola}
    \end{subfigure}
    \hfill
    \begin{subfigure}[b]{0.245\linewidth}
        \centering
        \includegraphics[width=\linewidth]{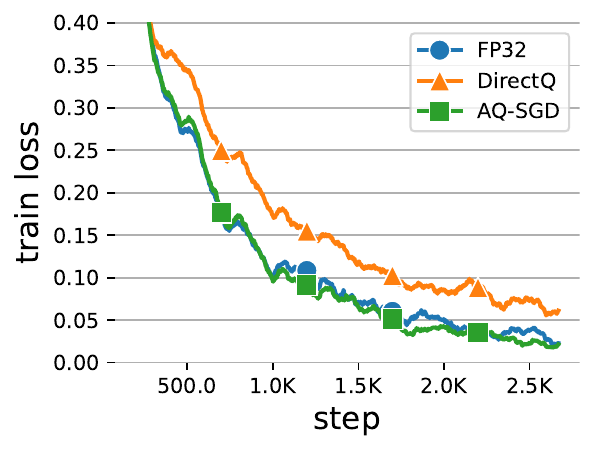}
        \caption{CoLA, DeBERTa-base}
        \label{fig:exp_cola_base}
    \end{subfigure}
    \hfill
    \begin{subfigure}[b]{0.245\linewidth}
        \centering
        \includegraphics[width=\linewidth]{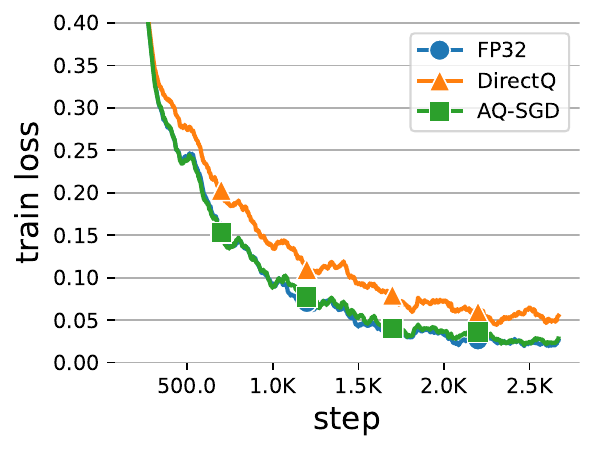}
        \caption{CoLA, DeBERTa-large}
        \label{fig:exp_cola_large}
    \end{subfigure}
    
    \caption{Convergence curves under different configurations.}
    \label{fig:exp_additional}
\end{figure}

\textbf{Number of Pipeline Stages.}
We first investigate the influence of the number of pipeline stages on convergence performance.
Intuitively, partitioning into more pipeline stages leads to more rounds of data compression and communication,resulting in a larger accumulated compression error.
The results of \Cref{fig:exp_n_cut_qnli,fig:exp_n_cut_cola} confirm this intuition.
Specifically, the direct quantization method works not bad when $K=2$, 
but its performance becomes unsatisfied when we further enlarge $K$.
In comparison, our approach can maintain similar convergence performance to \texttt{FP32}.

\textbf{Number of Bits in Communication.}
\Cref{fig:exp_n_bits_qnli,fig:exp_n_bits_cola} compare different methods with different numbers of bits in communication.
We observe that using more bits can improve the convergence performance but lead to higher communication overheads.
In general, our approach achieves better accuracy-efficiency trade-offs.

\textbf{Number of Bits for Previous Messages.}
We may find that storing all previous messages is space-intensive.
To reduce such requirements, we show that previous messages $m$ can be preserved with low precision.
We here perform quantization on $m$, where \texttt{m$z$} means that we use $z$ bits for previous messages.
\Cref{fig:exp_n_bits_cache_qnli,fig:exp_n_bits_cache_cola} show the results with different number of bits of the previous messages.
When only $2$ bits are used for the previous messages,
despite the fact that it is slightly worse than our default setting,
our approach is still significantly better than \texttt{DirectQ}.
And there is no significant performance drop when $8$ bits are used for the previous messages.

\textbf{Pre-trained Model Sizes.}
\Cref{fig:exp_cola_base,fig:exp_cola_large} show the results of the base and large version of DeBERTa.
Surprisingly, larger models seem to be more tolerant of errors from activation compression than smaller models.
One possible reason is that larger models usually use much smaller learning rates.
So the error of each iteration can be restricted to a smaller range.
Here, we use 2.0e-5 for the base model and 7e-6 for the large model, as suggested in the official repository of DeBERTa.

\subsection{Split Learning}
\label{exp:splitnn}

Split learning 
is a scenario of federated learning, where the client trains a shallow part of deep network (known as the cut layer) that accesses the training data, while the rest of the model is trained in a data center.
Clients and server need to exchange the activation and its gradients in the cut, where \algname can be adopted.  
We evaluated \algname on a split learning scenario where neither the input data nor its labels are shared with the server---the model is cut twice, one after the first resnet block and one before the last block to generate the prediction.
We evaluate \algname for split learning over Cifar10 and Cifar100 with the ResNet34 model.
We set 16 clients and use a Dirichlet distribution with concentration parameter 0.5 to synthesize non-identical datasets.
Following the previously established work of split learning,
in each communication round, we conduct local training for each client sequentially.
and each client will train 3 epochs with its local data.
We utilize SGD optimizer with momentum of 0.9, a batch size of 64, and a learning rate of 0.01.
We decay the learning rate to its 10\% for every 20 communication rounds.

The datasets are augmented with random cropping and flipping.
To adapt to random cropping, we do the same cropping operation on the retrieved previous message, and only update its non-cropped part.
To adapt to random flipping, we maintain another previous message copy for flipped images, and retrieve and update only the corresponding copy during training.

\begin{figure}
    \centering
    \begin{subfigure}[b]{0.245\linewidth}
        \centering
        \includegraphics[width=\linewidth]{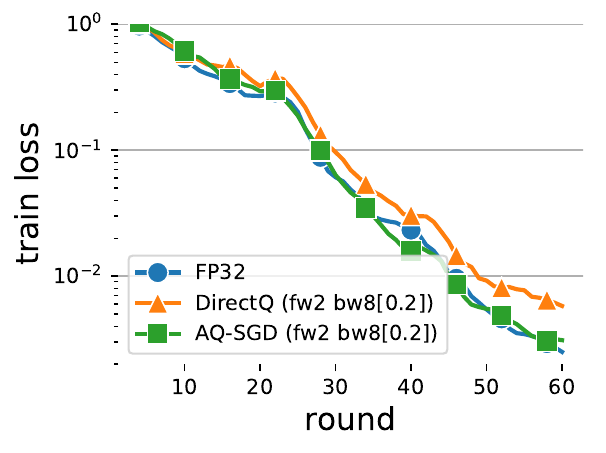}
        \caption{Cifar10, Train Loss}
    \end{subfigure}
    \hfill
    \begin{subfigure}[b]{0.245\linewidth}
        \centering
        \includegraphics[width=\linewidth]{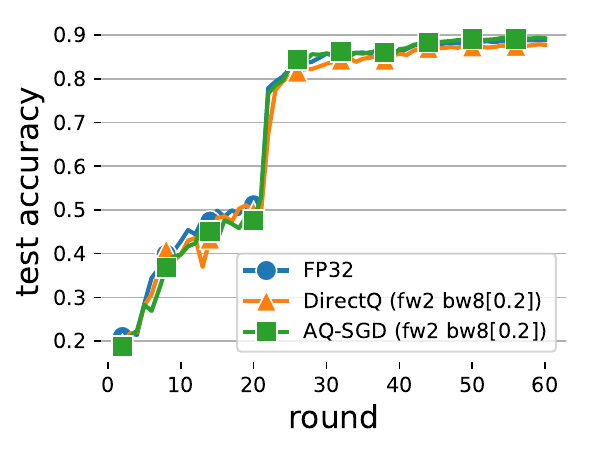}
        \caption{Cifar10, Test ACC}
    \end{subfigure}
    \hfill
    \begin{subfigure}[b]{0.245\linewidth}
        \centering
        \includegraphics[width=\linewidth]{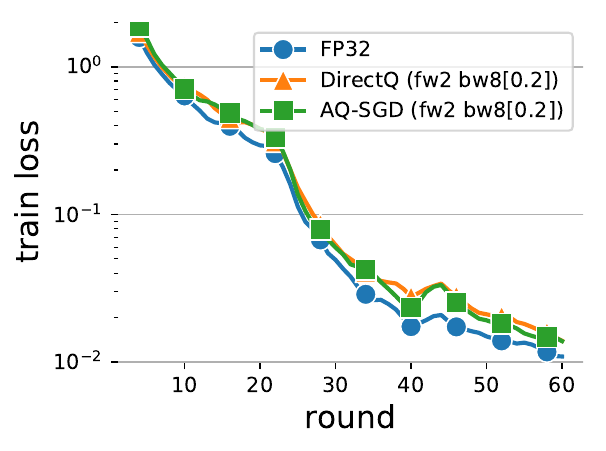}
        \caption{Cifar100, Train Loss}
    \end{subfigure}
    \hfill
    \begin{subfigure}[b]{0.245\linewidth}
        \centering
        \includegraphics[width=\linewidth]{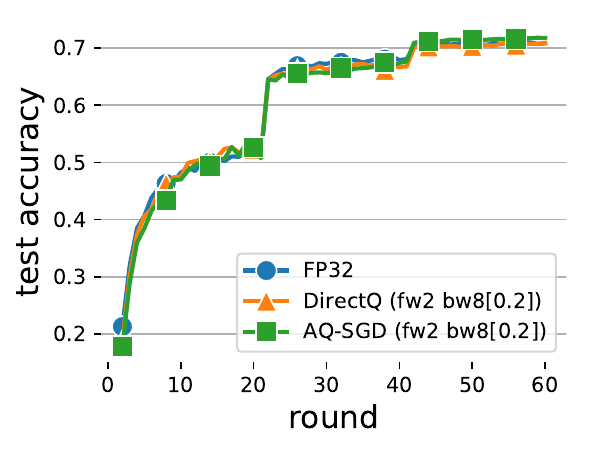}
        \caption{Cifar100, Test ACC}
    \end{subfigure}
       \caption{Results of split learning with ResNet34.}
       \label{fig:exp_split}
\end{figure}

Figure \ref{fig:exp_split} presents the results of split learning, 
where \texttt{fw2 bw8[0.2]} means that, 
for forward pass, we perform 2-bit quantization, and
for backward pass, we keep only the top 20\% gradients and then perform 8-bit quantization.
We can see that \algname transfers the activations in 2 bits while maintaining a performance similar to \texttt{FP32},
which indicates the effectiveness of \algname in improving the communication efficiency in the split learning scenario.
Furthermore, compared to \texttt{DirectQ},
\algname shows advantages in terms of both the convergence and generalization of the trained model.

{
% \color{blue}
\section{Case Study of Generation Results} \label{sec:case}

We here conduct case study to better understand the generation quality across different methods. 
All methods are fine-tuned on WikiText with the same seed. We use greedy decoding to generate texts.
For \texttt{DirectQ} and \algname, we use 4-bit in forward pass and 8-bit in backward pass.

We randomly pick documents from the test set of WikiText to construct the prompts and let the model complete the rest.
The generation results after fine-tuning are presented in Tables~\ref{tab:case} and \ref{tab:case2}.
For instance, in Cases~1 and 2, \algname produces the same results as \texttt{FP32};
in Case 3, although \algname does not generate the exact same output as \texttt{FP32}, it still expresses a similar meaning.
This indicates our proposed compression scheme has relatively little impact on the learning process.
In contrast, \texttt{DirectQ} may generate text that is not relevant to the context.
For example, in Case 1, we know that ``Matsuo Bashō was strongly influenced by Du Fu'', 
so we should expect the text below to show the fact how he was influenced, just like \texttt{FP32} and \algname did. 
But \texttt{DirectQ} fails to generate reasonable text based on the context.
Other inference results presented in Tables~\ref{tab:case} and \ref{tab:case2} lead to the same observation.

\begin{table}[t]
    \centering
    % \color{blue}
    % \captionsetup{labelfont={color=blue}}
    \caption{
    % \color{blue} 
    Case study of generation results.}
    \setlength{\tabcolsep}{2pt}
    \begin{tabular}{rp{0.8\linewidth}}
        \toprule
        \textbf{Case 1} &  \\ \midrule
        \textbf{Prompt:} & ... Matsuo Bashō, the greatest haiku poet, was also strongly influenced by Du Fu; in Oku no Hosomichi, his masterpiece, he cites the first two lines of A Spring View before a haiku as its introduction and also many of his other haiku have similar wording and themes. It is said that when he died \\
        \textbf{Gold:} & in Osaka during a long travel, a copy of Du Fu's poetry was found with him as one of a few precious items which he was able to carry around. \\
        \textbf{\texttt{FP32}}: & , his body was placed in a coffin with a Du Fu-like inscription.\\
        \textbf{\algname}: & , his body was placed in a coffin with a Du Fu-like inscription.\\
        \textbf{\texttt{DirectQ}}: & , his body was covered with a red cloth and his head was shaved.\\
        
        \midrule
        \textbf{Case 2} & \\ \midrule
        \textbf{Prompt}: & ... Peppiatt notes that the exhibition showed Bacon no longer needed sensationalist material to make an impact, and was now \\
        \textbf{Gold}: & capable of creating an intense emotional response through more subtle means ... \\
        \textbf{\texttt{FP32}}: & able to "make a serious contribution to the art of painting" \\
        \textbf{\algname}: & able to "make a serious contribution to the art of painting" \\
        \textbf{\texttt{DirectQ}}: & able to "make a serious statement about the state of the art" \\
        
        \midrule
        \textbf{Case 3} & \\ \midrule
        \textbf{Prompt}: & ... At this point, Dokhturov's column appeared behind the French line and joined the battle. The French were  outnumbered more than three to \\
        \textbf{Gold}: & one, assaulted in the front by Miloradovich's column, in the middle by Strik's and in the rear by Dokhturov. \\
        \textbf{\texttt{FP32}}: & one, but they held their ground and forced the Russians to retreat. \\
        \textbf{\algname}: & one, but they were able to hold their ground. \\
        \textbf{\texttt{DirectQ}}: & one, and the Russians were able to hold their ground. \\
        
        \midrule
        \textbf{Case 4} & \\ \midrule
        \textbf{Prompt}: & ... Between Jupiter and Fort Pierce, the storm knocked down power and telegraph lines. In the latter city, high waves washed out a portion of the causeway. In the 1980s, an elderly resident recalled that the storm was the \\
        \textbf{Gold}:  & most severe on record in Fort Pierce. \\
        \textbf{\texttt{FP32}}:  & worst storm he had ever seen. \\
        \textbf{\algname}:  & worst storm he had ever seen. \\
        \textbf{\texttt{DirectQ}}:  & worst he had ever experienced. \\

        \midrule
        \textbf{Case 5} & \\ \midrule
        \textbf{Prompt}: & ... Lesnar was born in Webster, South Dakota, on July 12, 1977. He was raised on a Webster dairy farm owned by his parents, Stephanie and Richard Lesnar. He is of German descent. He has two older brothers named Troy and Chad, and a younger sister named Brandi. At age 17, he joined the National Guard, where he \\
        \textbf{Gold}:  &  was assigned to an office job after his red-green colorblindness was deemed hazardous to his desire to work with explosives. \\
        \textbf{\texttt{FP32}}:  & served in the South Dakota National Guard. He was a member of the South Dakota National Guard for two years. \\
        \textbf{\algname}:  & served in the South Dakota National Guard. He was a member of the South Dakota National Guard for two years. \\
        \textbf{\texttt{DirectQ}}:  &  served in the South Dakota National Guard for two years. He was discharged in May 1991. \\
        
        \bottomrule
    \end{tabular}
    
    \label{tab:case}
\end{table}

\begin{table}[t]
    \centering
    % \color{blue}
    % \captionsetup{labelfont={color=blue}}
    \caption{
    % \color{blue} 
    Case study of generation results (more).}
    \setlength{\tabcolsep}{2pt}
    \begin{tabular}{rp{0.8\linewidth}}
        \toprule
        \textbf{Case 6} & \\ \midrule
        \textbf{Prompt}: & ... Mega placed the game at \# 7 in their "Top Mega Drive Games of All Time" list. MegaTech magazine praised the game's graphics, but criticized the game's easy difficulty level, \\
        \textbf{Gold}:  &  explaining simply that "the graphics are excellent, but the game is easy to complete." \\
        \textbf{\texttt{FP32}}:  &  stating that "the game is too easy to be enjoyable". \\
        \textbf{\algname}:  &  stating that it was "too easy". \\
        \textbf{\texttt{DirectQ}}:  &  which was criticized for being too easy. \\
        
        \midrule
        \textbf{Case 7} & \\ \midrule
        \textbf{Prompt}: & ... "Obviously, he has great rapport and communication ability with his team." Yahoo! Sports compared Stevens to legendary coach John Wooden writing " \\
        \textbf{Gold}:  &  Brad Stevens is winning at Butler the Wooden way – calm and composed on the sideline."  \\
        \textbf{\texttt{FP32}}:  &  He's a coach who has been around the block, and he's not afraid to tell you what he thinks."  \\
        \textbf{\algname}:  & Stevens is a coach who has been around the game for a long time, and he has a knack for getting the most out of his players." \\
        \textbf{\texttt{DirectQ}}:  &  Stevens is a coach who has been able to mold a team into a championship contender. He has a knack for getting the most out of his players, and he has a knack for getting the most out of his players' personalities." \\
        
        \midrule
        \textbf{Case 8} & \\ \midrule
        \textbf{Prompt}: & ... The Gore's social circle in England revolved around his good friend Rufus King, who \\
        \textbf{Gold}:  &   was appointed Ambassador to Great Britain in 1796, along with other Massachusetts expatriates. \\
        \textbf{\texttt{FP32}}:  &   was a member of the Royal Society.\\
        \textbf{\algname}:  &   was a member of the Royal Society. \\
        \textbf{\texttt{DirectQ}}:  &   was a member of the Royal Society. \\

        \midrule
        \textbf{Case 9} & \\ \midrule
        \textbf{Prompt}: & ... The city has 896 barangays that are known by sequential numbers instead of names. These barangays are further group into \\
        \textbf{Gold}:  &    100 zones for administrative and municipal purposes. \\
        \textbf{\texttt{FP32}}:  &    districts, which are further subdivided into wards.\\
        \textbf{\algname}:  &  districts, which are further subdivided into wards. \\
        \textbf{\texttt{DirectQ}}:  &   districts, which are numbered according to the number of barangays in the district.  \\

        \midrule
        \textbf{Case 10} & \\ \midrule
        \textbf{Prompt}: &  ... The National Spiritual Assembly of the Bahá 'ís of the Philippines, the governing body of the Filipino Bahá'í community, is \\
        \textbf{Gold}:  &     headquartered near Manila's eastern border with Makati. \\
        \textbf{\texttt{FP32}}:  &     headquartered in the city. \\
        \textbf{\algname}:  &   headquartered in the city. \\
        \textbf{\texttt{DirectQ}}:  &   headquartered in the city.  \\
        
        \bottomrule
    \end{tabular}
    
    \label{tab:case2}
\end{table}

}

\end{document}